\documentclass[letterpaper, 10 pt, conference]{ieeeconf}  

\IEEEoverridecommandlockouts                              

\usepackage{comment}
\usepackage{tcolorbox}
\usepackage[normalem]{ulem}
\usepackage{balance}
\usepackage{floatrow}
\usepackage{comment}
\usepackage{hyperref}
\usepackage{xr-hyper} 
\hypersetup{hypertexnames=false}
\usepackage{tcolorbox}
\usepackage{dsfont}
\usepackage[OT1]{fontenc}

\usepackage{cite}
\usepackage{amsmath,amssymb,amsfonts}
\usepackage[algo2e,ruled,vlined,linesnumbered]{algorithm2e}
\usepackage{algorithmicx}
\usepackage{graphicx,wrapfig,lipsum}
\usepackage{textcomp}
\usepackage{xcolor}

\usepackage{tcolorbox}
\usepackage{comment}
\usepackage{tikz}
\usetikzlibrary{arrows.meta,positioning}
\usepackage[normalem]{ulem}
\usepackage{balance}
\usepackage{floatrow}

\usepackage{amsthm}
\makeatletter
\let\savedmaketitle\maketitle
\let\saved@maketitle\@maketitle
\let\savedtitle\title
\let\savedauthor\author
\makeatother
\newcommand*{\addFileDependency}[1]{%
  \typeout{(#1)}%
  \@addtofilelist{#1}%
  \IfFileExists{#1}{}{\typeout{No file #1.}}%
}

\makeatother

\newfloatcommand{capbtabbox}{table}[][\FBwidth]

\newtheorem{theorem}{Theorem}[section] 

\newtheorem{definition}[theorem]{Definition}

\usepackage{booktabs}
\usepackage{url}

\let\labelindent\relax
\let\labelindent\relax
\usepackage{enumitem}
\setlist[description]{leftmargin=\parindent,labelindent=\parindent}

\def\BibTeX{{\rm B\kern-.05em{\sc i\kern-.025em b}\kern-.08em
    T\kern-.1667em\lower.7ex\hbox{E}\kern-.125emX}}

\title{\LARGE \bf Softness is Something You also Need! \\ Why Compliance Helps Control: \\ Information Metrics of a Passive Walker with Compliant Ankle
}

\author{Fabio Bonsignorio$^{1}$, Elisabetta Barletta, Sorin Dragomir, Francesco Esposito$^{2}$ and Enrica Zereik$^{3}$
\thanks{$^{1}$Fabio Bonsignorio is with University of Zagreb Faculty of Electrical Engineering and Computing, AIFORS LAB, 10000, Zagreb,  Croatia and Heron Robots, 16121, Genoa, Italy.
        {\tt\small fabio.bonsignorio@fer.hr}}%
        \thanks{$^{2}$Elisabetta Barletta, Sorin Dragomir and Francesco Esposito are with the School of Engineering of the University of Basilicata, 85100, Potenza, Italy.
        {\tt\small \{elisabetta.barletta, sorin.dragomir, f.esposito\}@unibas.it}}%
\thanks{$^{3}$Enrica Zereik is with the Institute of Marine Engineering of the National Research Council, 16149, Genoa, Italy
        {\tt\small enrica.zereik@cnr.it}}%
}

\begin{document}

\maketitle

\begin{abstract}
A great gap in adaptivity, flexibility, and dexterity still exists between intelligent robots, whether they are made of rigid or (partially) compliant components, and their naturally evolved animal and vegetal counterparts. One reason could be of `paradigm' nature. On the one hand,  most robots (both rigid and soft) are still designed by following a `Cartesian' approach, with top-down control and a clear mind-body separation. On the other hand, the principles of organization of animals and plants rely on morphological computation (which can be regarded as a partial or total outsourcing of information processing to the agent's dynamics) and emergence of fit behaviors from loosely coupled - sometimes huge and complex - networks of embodied agents.
Designing artificial agents on the basis of the principles that we observe in biology is not an easy task. Modeling and control require radically new approaches. In this paper we briefly recall and explain some seminal mathematical results linking information metrics and dynamics of the agents, already published years ago; we integrate them with some recent methods developed by us and by the community, we highlight some brief remarks about their relevance and we hint to our current and near future work towards finding concrete and efficient ways for the modeling and control of completely or partially compliant artificial agents and for a better understanding of physical intelligence in nature. We present a technical note summarizing phase--space volume calculations for inverted-pendulum and passive-dynamic walking models, following our proposed framework. Starting from
the McGeer compass gait, we introduce knee--ankle compliance, foot--rocker effects, and a Ruina-style compass
gait with distributed leg inertia. Standard heel--strike maps are derived and used in the estimation of viable phase--space volumes $W(\gamma)$, based on a state--dependent
Monte--Carlo and Reproducing Kernel Hilbert Spaces methods. We show how the methods based on RKHS allow deeper and more subtle analysis.
\end{abstract}


\section{Background}
\label{sec:background}

This work builds on \cite{bonsignorio2007preliminary}, utilizing the findings from \cite{Touchette2003infotheory}.
In \cite{bonsignorio2007preliminary} it is argued that there are good reasons to believe that 'intelligence' (and 'meaning') are 'emerging' traits arising from the evolution of loosely coupled networks of intelligent 'embodied'  and 'situated' agents.
The author, in line with the approach proposed by Rolf Pfeifer in ``How the body shapes the way we think'' \cite{Pfeifer:2006:HowtheBodyShapes}, elaborates a preliminary quantitative comprehensive theory that aims to enable us to predict the interaction between the agents' `morphology', or mechanical structure and dynamics, and the emergence of `intelligence' and `meaning' in physical systems.

Specifically, a number of relations are derived with a objective to clarify:
\begin{enumerate}
\item The relationship between an embodied agent's information/computing capability and its dynamics (morphological computation)
\item The behavior of computing and information capacities in a multi-body agent system
\item How networks of embodied agents give rise to 'intelligence' and 'meaning'
\end{enumerate}

This work has a narrower scope.
Our objective is to elucidate the significance of softness, or compliance, in robotics and natural embodied agents and provide quantitative guidance for hybrid hard and soft bodied robots.
\section{Mind--Body Nexus Equations}
The main modeling issue is, in that line of thought, to link the information processing metrics of the `mind' of the agent to those of the body's system dynamics.

In~\cite{bonsignorio2007preliminary} and \cite{Bonsignorio:2009:Steps} the following expressions were derived: 
\begin{equation}
{K(X)} \overset{+}{\operatorname{\le}} \ln \frac{\Delta W_{closed}}{\Delta W_{\text{open}}^{\max}
}
\label{eq:bI}\end{equation}

\begin{equation}
{\Delta H_N}+ \sum\limits_{i=1}^n {{\Delta H}_i}  - \Delta I \le I(X;C)
\label{eq:bII}\end{equation}

\begin{equation}
{K(X)} = {\Delta H_N}+ \sum\limits_{i=1}^n {{\Delta H}_i}  - \Delta I 
\label{eq:bIII}\end{equation}

\begin{equation}
{\Delta H_N} = \ln \frac{\Delta \Omega _{closed}}{\Delta  \Omega_{\text{open}}^{\max}
} + \Delta I
\label{eq:bIV}\end{equation}

Equation \eqref{eq:bI} follows in particular from:

\begin{equation}
H_{Shannon} = -\frac{ 1}{k_b} H_{Boltzman} =\frac{ 1}{k_b}\ln W
\label{eq:b5}\end{equation}
For details see~\cite{bonsignorio2007preliminary}.
The underlying assumptions are discussed in the next section.
\section{Discussion of equation \texorpdfstring{\eqref{eq:b5}}{5}: Meaning of \texorpdfstring{$W$}{W}}
The concept of entropy comes from statistical mechanics, but in particular if we follow an axiomatic probabilistic instead of a frequentistic approach, it can be generalized to a large class of systems including the biological ones, see~\cite{Baloghelal:2020:genentropies,Patcheretal:2024:entropyirr}.
Entropy allows to connect a large number of `micro' variables to a much smaller number of observed variables. 
In this case we can deduct the expected length of the control program of a dynamical system.
Equation \eqref{eq:b5} holds when all Shannon-Kolgomorov (S-K) four axioms \eqref{eq:bI}-\eqref{eq:bIV} are valid and the underlying dynamics is then estensive. This is true if we look at the possible paths in the phase--space of our systems. The S-K 4th axyom is satisfied as the trajectories of the system in the phase--space only depend on its dynamical equations. If we had for example hysteresis interactions among the paths (in a quantum) or any kind of history dependent behaviors the form of the relation $S(W)$, $S$ for thermodinamic entropy,  and thus of $H(W)$, $H$ for Shannon Entropy, would change, see ~\cite{Baloghelal:2020:genentropies}.
A more detailed discussion can be found in Supplementary Information in 
Supplementary Section~\ref{sec:wcalculation}. 

\section{Passive Walkers}
\begin{figure}[t]
\centering
\includegraphics[width=0.55\linewidth]{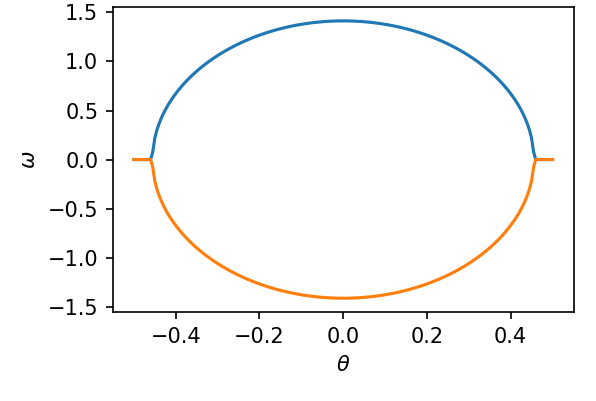}
\caption{Phase--space admissible region (passive leg).}
\label{fig:phase_space}
\end{figure}

\begin{figure}[t]
\centering
\includegraphics[height=4.25cm, width=0.5
\linewidth]{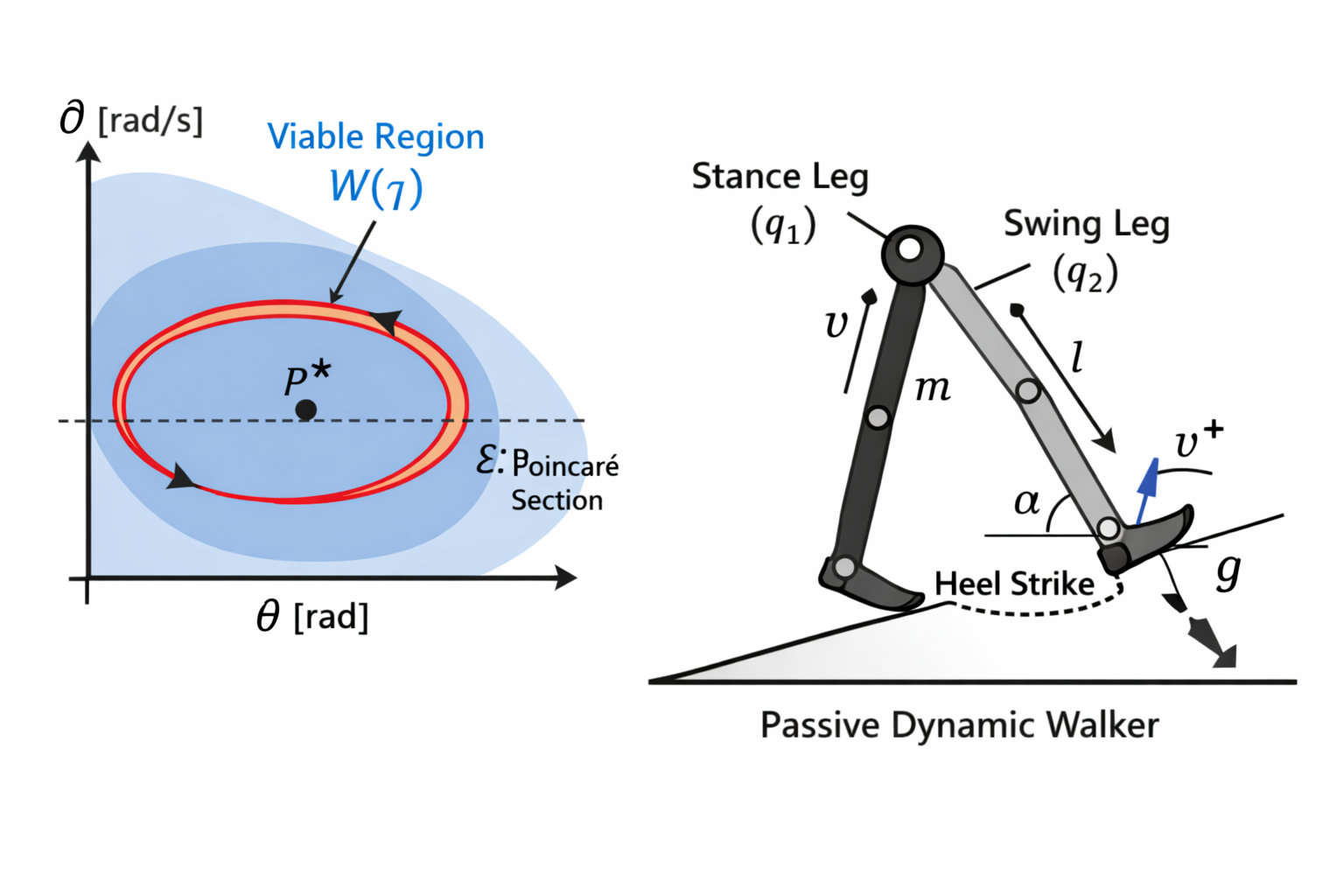}
\centering
\includegraphics[height=4.25cm, width=0.3
\linewidth]{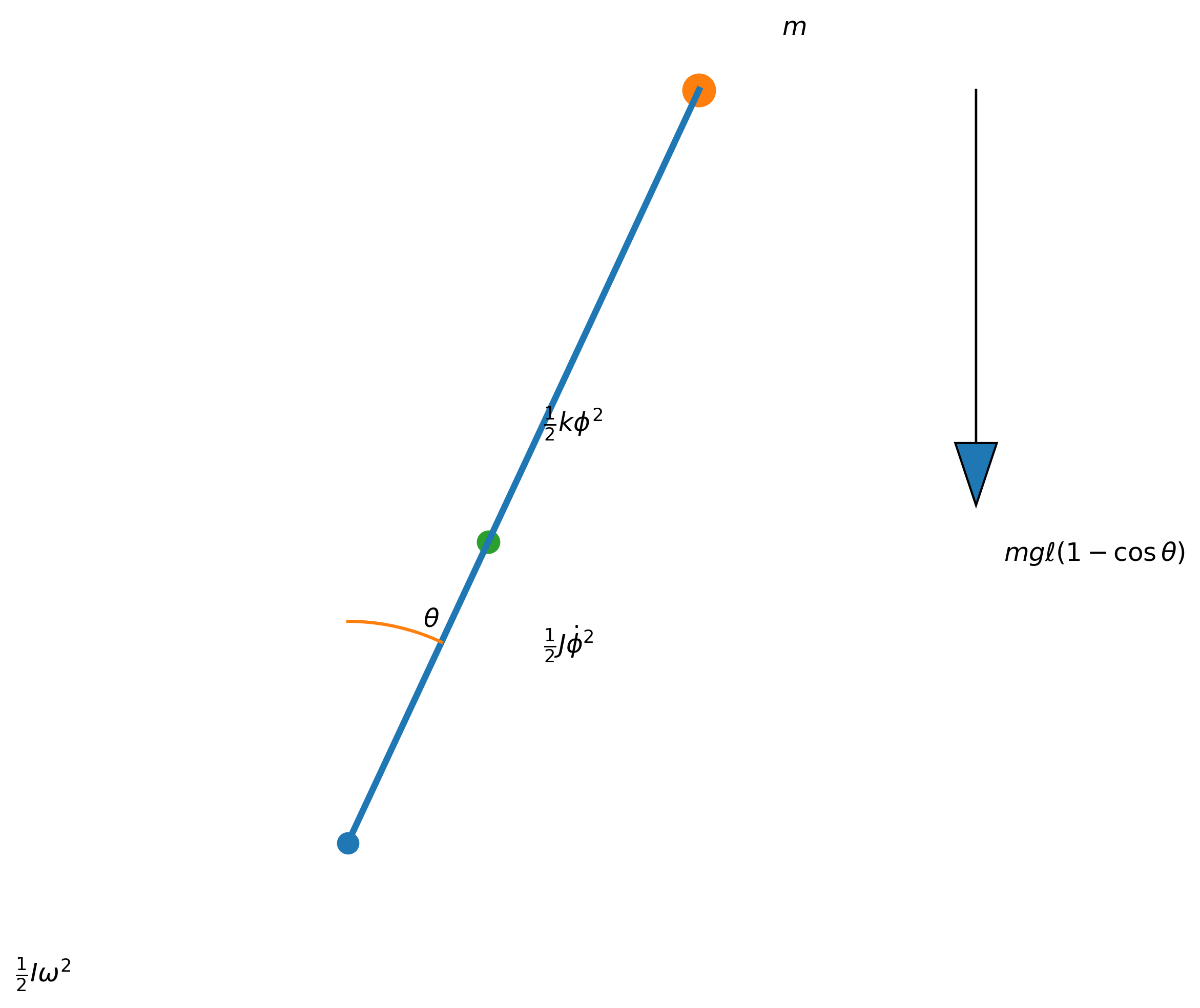}
\caption{Passive Walkers in the phase-space. Left: $\gamma$ is the angle of the slope for a pure passive walker. Center: Illustrative Picture. Right: Our Simplified Model}
\label{fig:passive_walker}
\end{figure}

Passive dynamic walkers are a powerful tool to illustrate morphological computation: body geometry and gravity restrict the set of admissible motions so strongly that little or no control is required. Figures~\ref{fig:phase_space} and~\ref{fig:passive_walker} show a representation of the phase-space and of the main variables for a simple passive walker, respectively.  Following the suggested framework able to link phase--space volume $W$ to information and control complexity, we compute $W$ for progressively richer walking morphologies.

\subsection{Ruina / Garcia Simplest Walker}
Model assumptions: hip point mass, massless legs, point feet, plastic heel--strike. The standard heel--strike law (exact result) is expressed as
\begin{equation}
\omega^{+} = \cos(2\alpha)\,\omega^{-}
\end{equation}
and the admissible domain is represented by:
\begin{equation}
D=\left\{(\theta,\omega): |\theta|\le \alpha,\; \tfrac12 I\omega^2 + m g \ell (1-\cos\theta)\le E^{+}\right\}.
\end{equation}
Hence, the phase--space volume can be easily computed:
\begin{equation}
W = \int_{-\alpha}^{\alpha} 2\sqrt{\frac{2}{I}\left(E^{+}-m g \ell (1-\cos\theta)\right)}\,d\theta.
\end{equation}

\subsection{Knee--Ankle Compliance and Foot--Rocker Effects}
Adding compliance to the structure (extra coordinate $\phi$) we obtain
\begin{equation}
H = \tfrac12 I\omega^2 + m g \ell (1-\cos\theta) + \tfrac12 J\dot\phi^2 + \tfrac12 k\phi^2
\label{eq:compliantinvpend}
\end{equation}
with main inverted pendulum angle $\theta$, tip mass $m$, internal compliant joint $\phi$, torsional spring $k$, and secondary inertia $J$.
The admissible domain becomes four-dimensional and its
geometry can be estimated by Monte--Carlo sampling.
Compliance can modify both the geometry of the viable
state space and the probability measure induced over it.
In particular, internal elastic energy storage may redistribute
occupancy over the task variables without necessarily
contracting their projected support. Reduced collision losses,
on the other hand, may modify the viable domain of the full
hybrid dynamics. Rocker feet can similarly affect viability
by modifying the impact mechanics and the corresponding
post-impact continuation conditions.\\
Equation~\ref{eq:compliantinvpend} is obtained by augmenting the standard
inverted-pendulum energy of the passive walker with a
single compliant internal coordinate $\phi$.
The compliant subsystem is modeled as a linear torsional
spring of stiffness $k$ and inertia $J$, yielding the
quadratic energy contribution

\[
H_\eta=\frac12 k\phi^2+\frac12 J\dot\phi^2.
\]

Such a representation corresponds to the small-deflection
approximation commonly adopted in compliant multibody
systems and passive-dynamic locomotion models.

\section{Compass Gait with Uniform-Rod Legs}
For a walker with uniform-rod legs, let us define coordinates $q=[q_1,q_2]^T$, each corresponding to a leg. Then, the swing-foot position can be expressed as
\begin{equation}
p_{\mathrm{sw}}(q)=\big[\ell(\sin q_1-\sin q_2),\;\ell(\cos q_1-\cos q_2)\big]^T
\end{equation}
and the Jacobian matrix can be computed:
\begin{equation}
J=\begin{bmatrix}
\ell\cos q_1 & -\ell\cos q_2\\
-\ell\sin q_1 & \ell\sin q_2
\end{bmatrix}.
\end{equation}
The heel--strike velocity reset is formulated as:
\begin{equation}
\dot q^{+}=\left(I-M^{-1}J^T(JM^{-1}J^T)^{-1}J\right)\dot q^{-},
\end{equation}
followed by leg relabeling. Numeric example at $\alpha=0.3$:
\begin{equation}
A \approx \begin{bmatrix}
0.8198 & -0.0154\\
0.5149 & -0.0191
\end{bmatrix}.
\end{equation}

\subsection{Full State--Dependent Monte--Carlo Estimation}
\begin{figure}
\begin{floatrow}
\ffigbox{%
\includegraphics[width=\linewidth]{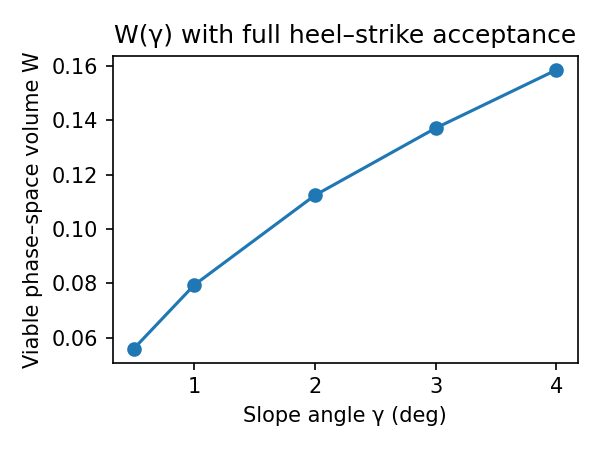}
}
{%
  \caption{$W(\gamma)$ with full heel--strike acceptance.}%
  \label{fig:wgamma}
}
\capbtabbox{%
\begin{tabular}{@{}cc@{}}
\toprule
$\gamma$ (deg) & $W(\gamma)$ \\ \midrule
0.5 & 0.056 \\
1.0 & 0.079 \\
2.0 & 0.112 \\
3.0 & 0.137 \\
4.0 & 0.159 \\ \bottomrule
\end{tabular}
}{%
    \caption{Representative viable phase--space volume values.}%
\label{tab:wgamma}
}
\end{floatrow}
\end{figure}

The post--impact viable set $D^{+}$ assumes:
\begin{equation}
q^{+}\approx[-\alpha,+\alpha]\pm dq,\qquad T^{+}\le E^{+}(\gamma).
\end{equation}
Monte--Carlo acceptance with exact heel--strike reset yields the representative values in Table~\ref{tab:wgamma}, also plotted in Figure~\ref{fig:wgamma}.

More details can be found in Supplementary Information Section \ref{sec:montecarlo}.
\subsection{Compliance, Phase--Space Geometry, and Informational Complexity}

As mentioned, the main modeling issue is to link the information processing metrics of the `mind' of the agent to those of the body's system dynamics.

Following~\cite{bonsignorio2007preliminary} and \cite{Bonsignorio:2009:Steps}, the complexity bound can be written as in equations (\ref{eq:bI}--\ref{eq:bIV}).
Equation~\eqref{eq:bI} follows in particular from the relation between Shannon and Boltzmann entropy, given in \eqref{eq:b5}:

\begin{equation}
H_{\text{Shannon}} = -\frac{1}{k_b} H_{\text{Boltzmann}} = \frac{1}{k_b} \ln W   \tag{5}
\end{equation}

where $W$ denotes the number (or phase--space measure) of admissible microstates of the system.

\subsection{Meaning of \texorpdfstring{$W$}{W} in Passive Dynamic Walking}

In passive dynamic walkers, the admissible configuration set is defined by an energy constraint in the phase--space.

For a simple inverted pendulum walker

\begin{equation}
D(E) = \{ (\theta,\omega) : H_x(\theta,\omega) \le E \}
\end{equation}

the Hamiltonian

\begin{equation}
H_x = \tfrac{1}{2} I \omega^2 + m g \ell (1 - \cos\theta)
\end{equation}
corresponds to the
standard energy function of a simple pendulum
\cite{goldstein, arnold}.

The viable phase--space volume is therefore:

\begin{equation}
W(E) = \int_{D(E)} d\theta \, d\omega
\end{equation}

Under a uniform occupancy model, $W(E)$ determines a
Boltzmann-type support entropy associated with the
admissible set. In the more general nonuniform case
considered below, however, scalar support volume alone does
not determine the distributional complexity; the probability
measure induced over the viable set must also be specified.

\subsection{Effect of Compliance: Support Geometry and
Induced Measure}

Consider the addition of compliant degrees of freedom,
\begin{equation}
z=(x,\eta),
\qquad
x=(\theta,\omega),
\qquad
\eta=(\phi,\dot{\phi}).
\end{equation}
For analytical tractability, assume the separable Hamiltonian
\begin{equation}
H(z)=H_x(x)+H_\eta(\eta),
\qquad
H_\eta(\eta)\geq0,
\end{equation}
where
\begin{equation}
H_\eta(\eta)
=
\frac{1}{2}k\phi^2+
\frac{1}{2}J\dot{\phi}^{\,2}.
\end{equation}
The rigid admissible set is
\begin{equation}
D_r(E)=\{x:H_x(x)\leq E\},
\end{equation}
whereas the compliant admissible set is
\begin{equation}
D_c(E)=
\{(x,\eta):H_x(x)+H_\eta(\eta)\leq E\}.
\end{equation}

Let
\begin{equation}
\Pi_xD_c(E)
=
\{x:\exists\eta\ {\rm such\ that}\ 
H_x(x)+H_\eta(\eta)\leq E\}
\end{equation}
denote the projection onto the task variables.

Since $H_\eta(\eta)\geq0$,
\begin{equation}
\Pi_xD_c(E)\subseteq D_r(E).
\end{equation}
However, the compliant subsystem possesses the admissible
zero-energy equilibrium
\begin{equation}
\eta_0=(0,0),
\qquad
H_\eta(\eta_0)=0.
\end{equation}
Therefore, for every $x\in D_r(E)$, choosing
$\eta=\eta_0$ gives
\begin{equation}
H_x(x)+H_\eta(\eta_0)=H_x(x)\leq E,
\end{equation}
and hence
\begin{equation}
\boxed{
\Pi_xD_c(E)=D_r(E).
}
\label{eq:projected_support_equivalence}
\end{equation}

Thus, in the separable model considered here, compliance
does not by itself contract the projected geometric support
of the task variables. Instead, the additional compliant
degrees of freedom modify how the available energy is
distributed over that support.

\subsection{Consequences for Informational Complexity}

The equality of projected supports does not imply equality
of the probability measures induced on those supports.
Denoting the rigid and compliant measures by $P_r$ and
$P_c$, respectively, one may have
\begin{equation}
\operatorname{supp}(P_r)
=
\operatorname{supp}(P_c)
=
D_r(E),
\end{equation}
while
\begin{equation}
P_r\neq P_c.
\end{equation}

Morphology can therefore influence locomotion complexity
through two conceptually distinct mechanisms: it may alter
the geometry of the viable set, or it may reshape the
probability distribution induced over that set. These effects
need not occur simultaneously.

Accordingly, the scalar viable phase--space volume $W$
should be interpreted as a zeroth-order descriptor of support
size rather than as a complete characterization of
informational complexity. Section~VII introduces an RKHS
description that incorporates the dynamically induced
probability measure and therefore distinguishes systems
having identical viable support but different statistical
organization.

\section{Montecarlo Analysis}
\label{sec:mcresults}
This section presents quantitative results based on Montecarlo simulation, demonstrating how the morphological properties of passive dynamic walkers affect the viable phase--space volume $W$ and the associated informational complexity measure $K(X)$.
More details can be found in the Supplementary Information section \ref{sec:montecarlo}.

\subsection{Simulation Setup}

We consider three progressively richer models of passive dynamic walking:

\begin{enumerate}
\item A rigid compass gait walker with two degrees of freedom $(\theta,\omega)$.
\item A compliant walker with an additional joint elasticity coordinate $(\phi,\dot\phi)$.
\item A compliant walker with distributed leg inertia and heel--strike dynamics.
\end{enumerate}

For each model, viable phase--space volumes were estimated using the Monte--Carlo method described in detail in  Supplementary Information section \ref{sec:montecarlo}. Samples were drawn uniformly in a bounding region and accepted if they satisfied both energy constraints and post--impact feasibility.

Kernel complexity was computed using a Gaussian kernel Gram matrix:

\begin{equation}
K(X) = \log \det(G + \lambda I).
\end{equation}

All simulations used $10^5$ samples drawn uniformly from the admissible energy-bounded region and a kernel bandwidth chosen according to Silverman's rule.
The reported averages are therefore geometric estimates of
the viable phase--space volume rather than dynamical
time averages.

\subsection{Effect of Compliance on the Task-State
Distribution}

The Monte--Carlo analysis illustrates the distinction between
geometric support and probability distribution discussed
above. Consider the local quadratic model
\begin{equation}
H_x=
\frac{1}{2}a\theta^2+
\frac{1}{2}b\omega^2,
\qquad
H_\eta=
\frac{1}{2}k\phi^2+
\frac{1}{2}J\dot{\phi}^{\,2}.
\end{equation}

For uniform sampling inside the two-dimensional rigid
energy ellipse,
\begin{equation}
\mathbb{E}_{r}[H_x]=\frac{E}{2}.
\end{equation}
For the compliant system, uniform sampling inside the full
four-dimensional energy ellipsoid instead gives
\begin{equation}
\mathbb{E}_{c}[H_x]=\frac{E}{3}.
\end{equation}
Hence
\begin{equation}
\boxed{
\mathbb{E}_{c}[H_x]
<
\mathbb{E}_{r}[H_x]
}
\end{equation}
despite
\begin{equation}
\Pi_xD_c(E)=D_r(E).
\end{equation}

Compliance therefore redistributes probability mass toward
lower-energy task states without necessarily reducing the
projected task-space support. The corresponding derivation
and numerical illustration are reported in Section~\ref{subsec:numerical_illustration}
and in the Supplementary Information.

\subsection{Impact of Heel--Strike Dynamics}

Including realistic heel--strike dynamics further reduced viable phase--space volume. The impact map introduces an instantaneous projection of the velocity state onto a lower--energy manifold.

For the Ruina--Garcia simplest walker, the heel--strike reset is
\begin{equation}
\omega^+ = \cos(2\alpha)\,\omega^- .
\end{equation}
For the value $\alpha=0.3\,\mathrm{rad}$ used in the numerical
example, the velocity-retention coefficient is
\begin{equation}
\rho \equiv \cos(2\alpha)
      = \cos(0.6)
      \simeq 0.8253 .
\end{equation}
Consequently, the kinetic energy associated with the angular
velocity is reduced at impact according to
\begin{equation}
\frac{T^+}{T^-}
=
\left(\frac{\omega^+}{\omega^-}\right)^2
=
\cos^2(2\alpha)
\simeq 0.681 .
\label{eq:impact_energy_ratio}
\end{equation}
Thus, approximately $32\%$ of the pre-impact kinetic energy
associated with this degree of freedom is lost at heel strike.

It is important to distinguish this energetic contraction from
the contraction of phase--space volume. In the reduced
$(\theta,\omega)$ representation, the reset map alone has
absolute Jacobian
\begin{equation}
\left|\det D\Delta\right|
=
|\cos(2\alpha)|
\simeq 0.825 .
\label{eq:impact_jacobian_ratio}
\end{equation}
Hence, the collision map alone contracts coordinate phase--space
area by approximately $17.5\%$. This geometric contraction must
be distinguished from the reduction in kinetic energy,
$T^+/T^-\simeq 0.681$, and from any change in the viable
phase--space volume $W(D_v)$. The latter additionally depends
on which post-impact states can successfully complete the
subsequent step and is therefore a property of the full hybrid
walking dynamics rather than of the reset map alone.

This shows that collision dynamics plays a significant role in shaping the accessible phase--space of passive locomotion systems.

\section{RKHS Geometry of the Viable State Distribution}
\label{sec:rkhs}

The previous sections characterized the viable behavior of passive
walking systems through Hamiltonian energy constraints and
projected phase--space geometry.
However, geometry alone does not completely determine the
computational burden associated with locomotion.

The crucial observation is that the walking dynamics does not merely
define a viable set of admissible states.
It also induces a probability measure over this set.
While different morphologies may possess identical geometric
support, they generally generate different probability
distributions over the corresponding viable trajectories.

Consequently, the complexity experienced by the controller should
not be regarded solely as a function of the size of the viable
region, but rather as a function of the probability measure
supported on that region.

This viewpoint naturally leads to a geometric-statistical
description based on reproducing kernel Hilbert spaces (RKHS),
where both the geometry of the viable manifold and the probability
measure induced by the dynamics can be represented within a common
mathematical framework.

\subsection{Geometry of the Viable State Space}

Let $ \mathcal X$ denote the state space of the walker.

For a given total energy budget $E$ the Hamiltonian

\begin{equation}
H:\mathcal X\rightarrow\mathbb R
\end{equation}

defines the viable domain

\begin{equation}
D(E)
=
\left\{
x\in\mathcal X:
H(x)\le E
\right\}.
\end{equation}

When heel--strike events are present,
only states whose reset map produces an admissible
post-impact configuration are retained.
Accordingly, the effective viable domain becomes

\begin{equation}
D_v(E)
\subseteq
D(E).
\end{equation}

The set
$ D_v(E)
$ is therefore determined jointly by

\begin{itemize}
\item energetic feasibility;
\item collision constraints;
\item impact consistency;
\item passive stability.
\end{itemize}

It represents the subset of the state space from which passive
walking can continue without external intervention.

Notice that this construction is purely geometric.
It specifies which states are admissible, but it does not specify
how frequently these states are visited during locomotion.

Consequently, $D_v(E) $ should be interpreted as the support of the dynamics rather than
their statistical description.

For later developments we denote by

\begin{equation}
W(D_v)
=
\int_{D_v} dV
\label{eq:viable_region_volume}
\end{equation}

the geometric volume of the viable region with respect to the
chosen reference measure.

Although this quantity provides an intrinsic measure of the size
of the viable domain, it does not capture how trajectories are
distributed inside it.

\subsection{Dynamically Induced Probability Measure}

The walking dynamics generates trajectories
$ x(t)\in D_v$ which induce a probability measure $P$ over the viable domain.

Instead of regarding every viable state as equally relevant,
the probability measure weighs each region according to the
frequency with which it is visited by the dynamics.

For any measurable subset $ A \subseteq D_v $ the induced measure is defined as

\begin{equation}
P(A)
=
\lim_{T\rightarrow\infty}
\frac{1}{T}
\int_0^T
\mathbf 1_A(x(t))
\,dt,
\end{equation}

whenever the limit exists, and where $\mathbf 1_A(x(t))$ denotes the indicator function of the set $A$.

Equivalently, if the dynamics admits an invariant density

$
\rho(x)
$ then

\begin{equation}
dP(x)
=
\rho(x)\,dV(x),
\end{equation}

with

\begin{equation}
\int_{D_v}
\rho(x)
\,dV(x)
=
1.
\end{equation}

The geometric volume $
W(D_v)$ and the probability measure $P$ are fundamentally different objects.

\emph{The former characterizes the support of the viable dynamics, while the latter characterizes how the dynamics populate that support.}

This distinction is particularly important when comparing rigid
and compliant walkers.

For the separable Hamiltonian

\begin{equation}
H(x,\eta)
=
H_x(x)
+
H_\eta(\eta),
\end{equation}

with

\begin{equation}
H_\eta(\eta)\ge0,
\end{equation}

the projection of the compliant viable set onto the task
variables may coincide with the rigid support,

\begin{equation}
\Pi_xD_c(E)
=
D_r(E),
\end{equation}

even though the corresponding induced measures differ.

Indeed, numerical simulations presented later show that
uniform sampling inside the full compliant energy ellipsoid
produces a projected distribution that is substantially more
concentrated around low-energy task states than the rigid
distribution (refer to Figure \ref{fig:task_state_distributions}).

Consequently, $
P_{\mathrm{compliant}}
\neq
P_{\mathrm{rigid}},
$ despite having essentially identical geometric support.

As already suggested, this observation strongly motivates the exploitation of the
RKHS formulation.

Rather than measuring only the size of the viable region,
we seek to quantify the complexity of the probability measure
supported on that region.
In this perspective,

\begin{tcolorbox}[colframe=blue, colback=blue!20]
\emph{Geometry determines where locomotion is possible,}

\emph{Dynamics determine how locomotion is distributed,}

\emph{RKHS quantify the complexity of that distribution.}
\end{tcolorbox}

\subsection{RKHS Embedding of the Viable State Distribution}

The previous subsection established that the relevant object is
not only the viable state space itself, but also the probability
measure induced by the walking dynamics.

This naturally leads to the use of reproducing kernel Hilbert
spaces, where probability measures can be represented as geometric
objects through their kernel mean embeddings.

Let $ k:\mathcal X\times\mathcal X
\rightarrow
\mathbb R
$ be a continuous positive-definite kernel and let $\mathcal H_k $ denote the associated reproducing kernel Hilbert space.

The canonical feature map is

\begin{equation}
\Phi(x)
=
k(\cdot,x).
\label{eq:feature_map}
\end{equation}

Given the dynamically induced probability measure $P$ introduced in the previous subsection, its RKHS representation is
defined through the kernel mean embedding

\begin{equation}
\mu_P
=
\int_{\mathcal X}
k(\cdot,x)
\,dP(x)
\label{eq:kernel_mean_embedding}
\end{equation}

whenever the integral exists in the Bochner sense. This representation can be interpreted as the RKHS-valued expectation of the feature map $k(\cdot,x)$.

The kernel mean embedding provides a complete representation of
the probability measure whenever the kernel is characteristic.
Consequently,

\begin{equation}
P_1=P_2
\Longleftrightarrow
\mu_{P_1}
=
\mu_{P_2},
\end{equation}

allowing different walker morphologies to be compared directly in
feature space.

Unlike geometric volume, which only characterizes the support of
the viable region, the kernel mean embedding captures both the
support and the probability mass distributed over it.

Therefore two walkers possessing identical viable domains may
still produce different RKHS representations whenever the induced
probability measures differ.

This situation naturally arises in compliant passive walkers.
Although the projected viable domains may coincide,

\begin{equation}
\Pi_xD_c(E)
=
D_r(E),
\end{equation}

their induced measures satisfy

\begin{equation}
P_{\mathrm c}
\neq
P_{\mathrm r},
\end{equation}

leading to distinct kernel mean embeddings

\begin{equation}
\mu_c
\neq
\mu_r.
\end{equation}

The computational complexity experienced by the controller is
therefore associated with the geometry of these probability
embeddings rather than with the size of the viable domain alone.

\subsection{Pullback Geometry}

The RKHS embedding induces a natural differential geometry on the
viable state distribution.

Following the coherent-state construction introduced in
Section~IV, every state $ x\in D_v
$ is mapped onto the projective Hilbert space through

\begin{equation}
\mathcal K:
D_v
\rightarrow
\mathbb P(\mathcal H_k),
\qquad
x
\mapsto
[k(\cdot,x)].
\end{equation}

The projective Hilbert space is endowed with the
Fubini--Study metric $g_{FS}$,  \cite{tozzi2026fubinistudy}.

Pulling this metric back through $ \mathcal K $ induces the Riemannian metric

\begin{equation}
g_k
=
\mathcal K^*
g_{FS}
\end{equation}

on the viable state manifold.

The corresponding infinitesimal distance is

\begin{equation}
ds^2
=
g_{ij}(x)
dx^idx^j.
\end{equation}

Unlike the Euclidean metric of the original coordinates, this
metric measures distinguishability in feature space.

Two states that are close in Euclidean coordinates may become
well separated whenever they induce substantially different kernel
responses.

Conversely, dynamically similar states become naturally clustered,
independently of the coordinate system used to describe the
walker.

The pullback metric therefore defines an intrinsic geometry of the
viable dynamics rather than of the state coordinates.

The corresponding Riemannian volume form is

\begin{equation}
dV_k
=
\sqrt{\det g_k(x)}
\,dx.
\end{equation}

Integrating this volume form over the viable domain yields the
kernel-induced geometric volume

\begin{equation}
W_k
=
\int_{D_v}
dV_k.
\end{equation}

Notice, however, that $
W_k
$ should not generally be interpreted as the physical phase-space
volume.

Instead, it measures the geometric size of the viable region under
the similarity notion induced by the kernel.

The equality $W_k = W_{\mathrm{phys}}$ holds only if the pullback metric coincides with the physical metric, $g_k = g_{\mathrm{phys}}$.

More generally, $
W_k
$ defines an intrinsic geometric measure associated with the RKHS
representation of viable locomotion.
This distinction is fundamental.

The geometric volume $
W_k
$ characterizes the manifold supporting the dynamics, whereas the probability measure $
P
$ characterizes how trajectories populate that manifold.

The next subsection combines these two ingredients by introducing
a covariance operator over the RKHS, from which an intrinsic
measure of distributional complexity is obtained.

\subsection{RKHS Distributional Complexity}
\label{subsec:rkhs_distributional_complexity}

The previous sections introduced the geometric volume $W(D_v)$ in Equation (~\ref{eq:viable_region_volume}) 
as a quantitative measure of the size of the viable state space.

This quantity provides a zeroth-order geometric descriptor
of the size of the viable state space and, under the
assumptions underlying Equation~\eqref{eq:b5} , can be related to a
support-based informational description of the walking task.

However, the geometric volume alone does not completely determine
the effective complexity experienced by the controller.

Indeed, two walkers may possess viable regions of identical
geometric size while exhibiting substantially different dynamical
behavior owing to different probability distributions over the
same viable manifold.

The volume therefore characterizes the geometry of the support,
whereas the induced probability measure characterizes how this
support is actually explored by the dynamics.

The natural extension of the geometric volume is thus obtained by
considering the pair $ (D_v,P)$, 
where \(P\) denotes the dynamically induced probability measure
introduced in the previous subsection.

Within the RKHS framework this pair is represented through its
kernel mean embedding and covariance operator.

From Equations (\ref{eq:feature_map}) and (\ref{eq:kernel_mean_embedding}) recall that $\Phi(x)=k(\cdot,x)$ denotes the feature map associated with the reproducing kernel, and that the kernel mean embedding of the viable dynamics is $\mu_P
=
\int
\Phi(x)\,dP(x)$. The centered covariance operator is then expressed as:
\begin{equation}
C_P
=
\int
(\Phi(x)-\mu_P)
\otimes
(\Phi(x)-\mu_P)
\,dP(x).
\end{equation}

This covariance operator describes how the probability measure is
distributed over the geometric directions defined by the RKHS.

If the dynamics occupies only a few independent directions, the
spectrum of \(C_P\) becomes concentrated. Conversely, broadly
distributed dynamics generate covariance operators with larger
effective rank.

This observation motivates the following definition.

\begin{definition}[RKHS Distributional Complexity]

Let \(C_P\) be the covariance operator associated with the
probability measure \(P\).

The RKHS distributional complexity is defined as

\begin{equation}
\boxed{
\mathcal C(P)
=
\log
{\det}_{F}
(I+\gamma C_P),
}
\end{equation}

where \(\gamma>0\) is a scaling parameter and
\(\det_F\) denotes the Fredholm determinant.

Equivalently,

\begin{equation}
\mathcal C(P)
=
\sum_i
\log
(1+\gamma\lambda_i),
\end{equation}

where \(\{\lambda_i\}\) are the eigenvalues of \(C_P\).

\end{definition}

The complexity introduced above naturally generalizes the
phase-space volume employed throughout the previous sections.

Indeed, if the viable dynamics populate the viable manifold
uniformly,

\begin{equation}
dP
=
\frac{1}{W(D_v)}
\,dV,
\end{equation}

implying that the probability distribution is completely determined by the
geometry of the support.

In this limiting situation, the probability measure is fixed
once the viable manifold and its reference volume measure are
specified. The covariance operator therefore depends only on
the geometry of the viable support as represented through the
chosen kernel. Importantly, this does not imply that the
complexity is a function of the scalar volume $W(D_v)$ alone:
domains having identical volume may possess different shapes,
metric structures, and kernel covariance spectra.

Accordingly, $W(D_v)$ should be interpreted as a
zeroth--order descriptor of viable-support size, whereas
$\mathcal{C}(P)$ provides a richer characterization of the
metric--measure structure of the viable dynamics.

\textbf{Theorem 7.2 (Generalized Complexity Principle).}
Let $P$ be a probability measure supported on the viable
manifold $D_v$, and let
\begin{equation}
\mathcal{C}(P)
=
\log\det_F\!\left(I+\gamma C_P\right)
\end{equation}
denote its RKHS distributional complexity.

Then the following statements hold.

\begin{enumerate}
\item If the viable dynamics populate $D_v$ uniformly,
\begin{equation}
dP(x)
=
\frac{1}{W(D_v)}\,dV(x),
\label{eq:uniform_viable_measure}
\end{equation}
then
\begin{equation}
\mathcal{C}(P)
=
\mathcal{F}_k(D_v,dV),
\label{eq:geometric_complexity}
\end{equation}
where $\mathcal{F}_k$ is a functional of the geometry of the
viable manifold and of the chosen kernel $k$.

Thus, under uniform occupancy, the distributional complexity
is completely determined by the geometry of the viable support,
but not, in general, by its scalar volume $W(D_v)$ alone.

\item For an arbitrary probability measure $P$,
\begin{equation}
\mathcal{C}(P)
=
\mathcal{F}_k(D_v,P),
\label{eq:general_distributional_complexity}
\end{equation}
and therefore depends jointly on the geometry of the viable
support and on the probability distribution induced over that
support by the dynamics.
\end{enumerate}

\emph{Proof.}
For the uniform measure,
\begin{equation}
dP(x)
=
\frac{1}{W(D_v)}\,dV(x),
\end{equation}
the kernel mean and covariance operator are
\begin{equation}
\mu_P
=
\frac{1}{W(D_v)}
\int_{D_v}\Phi(x)\,dV(x)
\end{equation}
and
\begin{equation}
C_P
=
\frac{1}{W(D_v)}
\int_{D_v}
\bigl(\Phi(x)-\mu_P\bigr)
\otimes
\bigl(\Phi(x)-\mu_P\bigr)
\,dV(x).
\end{equation}
Hence the spectrum of $C_P$, and consequently
$\mathcal{C}(P)$, is determined by the full geometry of
$D_v$ as represented through the kernel $k$.

In general, two viable domains may have equal scalar volume,
\begin{equation}
W(D_v^{(1)})=W(D_v^{(2)}),
\end{equation}
while possessing different shapes or metric structures.
Their kernel covariance spectra may therefore differ, implying
\begin{equation}
\mathcal{C}(P_1)\neq\mathcal{C}(P_2).
\end{equation}
Consequently, scalar phase--space volume alone is not
sufficient to determine RKHS distributional complexity.

For a nonuniform measure $P$, the covariance operator
additionally depends on how probability mass is distributed
within $D_v$. Therefore $\mathcal{C}(P)$ is a functional of
the metric--measure structure represented by $(D_v,P)$.
\hfill$\square$

\subsection{Numerical Illustration}
\label{subsec:numerical_illustration}

We now illustrate the distinction between geometric support,
kernel-induced volume, and RKHS distributional complexity by means
of a quadratic passive-walker model.

Let the task variables be
\begin{equation}
x=(\theta,\omega),
\end{equation}
and let the compliant internal variables be
\begin{equation}
\eta=(\phi,\dot{\phi}).
\end{equation}
We consider the quadratic energies
\begin{equation}
H_x(\theta,\omega)
=
\frac{1}{2}a\theta^2
+
\frac{1}{2}b\omega^2,
\end{equation}
and
\begin{equation}
H_\eta(\phi,\dot{\phi})
=
\frac{1}{2}k\phi^2
+
\frac{1}{2}J\dot{\phi}^2.
\end{equation}

The numerical parameters are
\begin{equation}
a=b=J=E=1,
\qquad
k=4.
\end{equation}

The rigid viable set is
\begin{equation}
D_r(E)
=
\left\{
x:
H_x(x)\le E
\right\},
\end{equation}
whereas the compliant viable set in the augmented state space is
\begin{equation}
D_c(E)
=
\left\{
(x,\eta):
H_x(x)+H_\eta(\eta)\le E
\right\}.
\end{equation}

\subsubsection{Support Geometry}

Since the internal equilibrium state $\eta=0$ is admissible, every task state satisfying
\begin{equation}
H_x(x)\le E
\end{equation}
admits at least one compliant realization. Therefore,
\begin{equation}
\Pi_x D_c(E)
=
D_r(E).
\end{equation}

Hence, for this separable quadratic model, compliance does not
reduce the support of the projected viable set. In particular, the
rigid and compliant projected task-space volumes are equal.

The exact Euclidean area of the rigid ellipse is
\begin{equation}
W_{\mathrm{Euc}}
=
\frac{2\pi E}{\sqrt{ab}}
=
2\pi
\approx
6.2832.
\end{equation}

A rejection-sampling Monte--Carlo estimate using
$N=3\times10^5$
samples gave
\begin{equation}
\widehat W_{\mathrm{Euc}}
=
6.2891,
\end{equation}
which is consistent with the analytical value.

\subsubsection{Dynamically Induced Task-State Distribution}

Although the projected supports coincide, the probability measures
induced on task space are different.

For the rigid walker, samples were drawn uniformly from the
two-dimensional energy ellipse. For the compliant walker, samples
were drawn uniformly from the four-dimensional energy ellipsoid
and subsequently projected onto the task variables.

The resulting rigid distribution is uniform over the task-space
ellipse, whereas the projected compliant distribution is more
concentrated around low-energy task states.

For uniform sampling inside a \(d\)-dimensional quadratic energy
ball, the expected total normalized energy is
\begin{equation}
\mathbb E[H]
=
\frac{d}{d+2}E.
\end{equation}

For the rigid two-dimensional model, all quadratic coordinates
belong to the task subsystem, and therefore
\begin{equation}
\mathbb E[H_x]_{\mathrm{rigid}}
=
\frac{2}{2+2}E
=
\frac{E}{2}.
\end{equation}

For the compliant four-dimensional model,
\begin{equation}
\mathbb E[H_x]
=
\frac{4}{4+2}E
=
\frac{2E}{3}.
\end{equation}
Since two of the
four quadratic coordinates belong to the task subsystem, the energy fraction received by such subsystem is $\frac{2}{4}$, implying:

\begin{equation}
\mathbb E[H_x]_{\mathrm{compliant}}
=
\frac{1}{2}E[H_x]
=
\frac{E}{3}.
\end{equation}

The simulations gave
\begin{equation}
\widehat{\mathbb E}[H_x]_{\mathrm{rigid}}
=
0.5004,
\end{equation}
and
\begin{equation}
\widehat{\mathbb E}[H_x]_{\mathrm{compliant}}
=
0.3328,
\end{equation}
in close agreement with the theoretical values $\frac{1}{2}$ and $\frac{1}{3}$.

\begin{figure}[t]
\centering
\includegraphics[width=\columnwidth]
{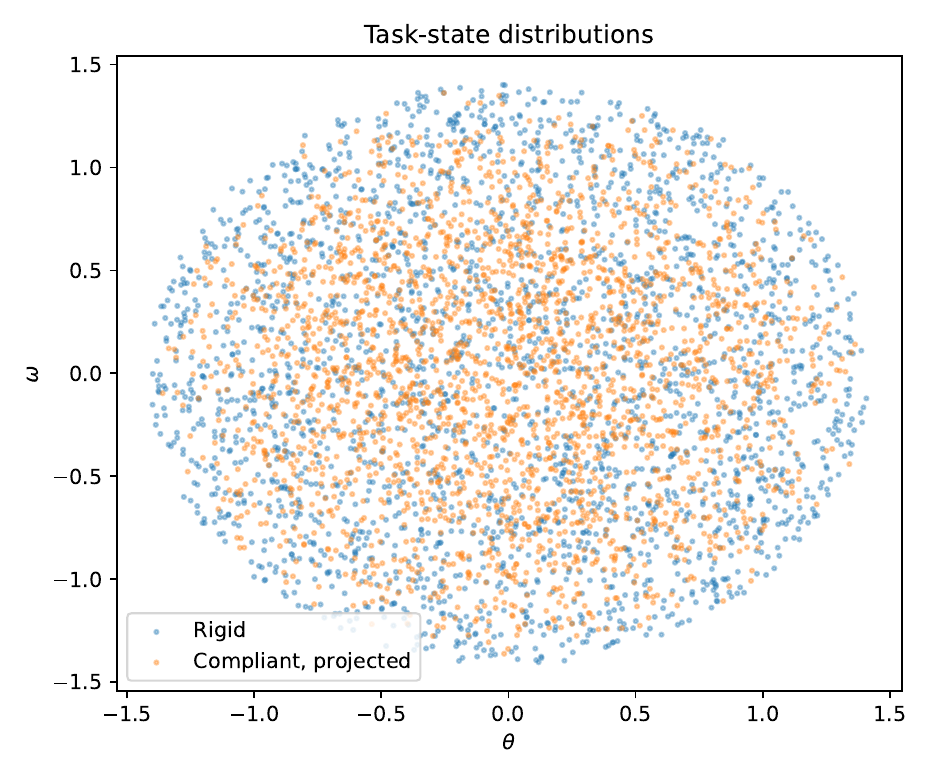}
\caption{Rigid and projected compliant task-state samples. The two
distributions have the same support, but the projected compliant
distribution is more concentrated around low-energy states.}
\label{fig:task_state_distributions}
\end{figure}

\begin{figure}[t]
\centering
\includegraphics[width=\columnwidth]
{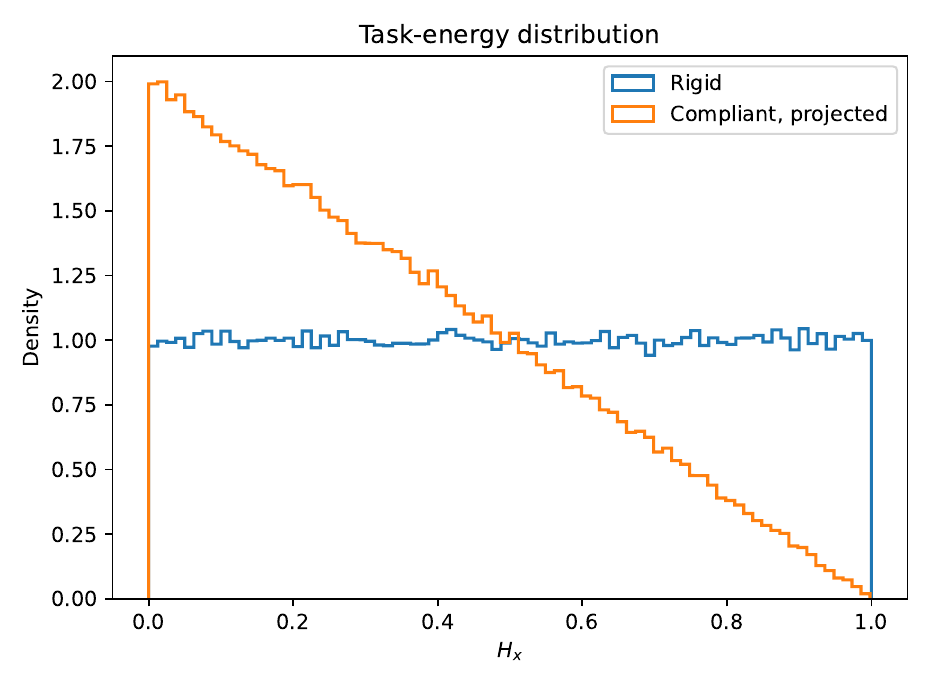}
\caption{Task-energy distributions for rigid and projected
compliant samples. Uniform sampling in the full compliant energy
ellipsoid shifts probability mass toward lower values of
\(H_x\).}
\label{fig:task_energy_distribution}
\end{figure}

Figure~\ref{fig:task_state_distributions} shows that the two
distributions have the same support, while the compliant
projection is more densely concentrated near the origin.
Figure~\ref{fig:task_energy_distribution} shows the corresponding
shift toward lower task energy.

\subsubsection{Kernel-Induced Pullback Volume}

To illustrate the pullback construction, we used the
inhomogeneous polynomial kernel
\begin{equation}
K(x,y)
=
(c+x^\top y)^p,
\end{equation}
with
\[
c=1,
\qquad
p=2.
\]

For this kernel, the Fubini--Study pullback metric on the
two-dimensional task space has volume density
\begin{equation}
\sqrt{\det g_K(x)}
=
\frac{p\sqrt{c}}
{\left(c+\lVert x\rVert^2\right)^{3/2}}.
\end{equation}

The corresponding kernel-induced volume was estimated as
\begin{equation}
\widehat W_K
=
\frac{\operatorname{Vol}(B)}{N}
\sum_{i=1}^{N}
\mathbf 1_{D_r}(x_i)
\sqrt{\det g_K(x_i)}.
\end{equation}

The numerical result was
\begin{equation}
\widehat W_K
=
5.3171.
\end{equation}

Since the rigid and projected compliant supports coincide and the
same kernel-induced metric is used in both cases, their pullback
volumes also coincide.

This confirms that pullback volume remains a geometric quantity:
it distinguishes geometries, but it does not by itself distinguish
different probability measures supported on the same viable set.

\subsubsection{Empirical RKHS Distributional Complexity}

To quantify the difference between the induced probability
measures, we computed the empirical RKHS distributional complexity.

Given $ N_s=300 $
task-state samples, the Gaussian Gram matrix was constructed as
\begin{equation}
G_{ij}
=
\exp
\left(
-\frac{\lVert x_i-x_j\rVert^2}{2\sigma^2}
\right).
\end{equation}

The bandwidth \(\sigma\) was selected once from the pooled rigid
and compliant samples and then kept fixed for both models.

Let $\mathbf 1_N \in \mathds{R}^N$ denote the N-dimensional vector of ones and define
\begin{equation}
H_N
=
I_N
-
\frac{1}{N}
\mathbf 1_N\mathbf 1_N^\top
\end{equation}
as the centering matrix. The empirical complexity was evaluated as
\begin{equation}
\widehat{\mathcal C}_\lambda
=
\log\det
\left(
I_N
+
\frac{1}{N\lambda}
H_NGH_N
\right),
\end{equation}
with
\[
\lambda=10^{-2}.
\]

The experiment was repeated over 20 independent realizations. The resulting mean values were
\[
\widehat{\mathcal C}_\lambda^{\mathrm{rigid}}
=
11.935,
\]
and
\[
\widehat{\mathcal C}_\lambda^{\mathrm{compliant}}
=
9.865.
\]

Thus,
\begin{equation}
\widehat{\mathcal C}_\lambda^{\mathrm{compliant}}
<
\widehat{\mathcal C}_\lambda^{\mathrm{rigid}},
\end{equation}
even though
\begin{equation}
\Pi_x D_c(E)
=
D_r(E).
\end{equation}

The reduction is therefore distributional rather than
support-based. The compliant morphology redistributes probability
mass toward a smaller number of effective feature-space
directions, thereby reducing the spectral spread of the centered
kernel covariance operator.

\begin{figure}[t]
\centering
\includegraphics[width=\columnwidth]
{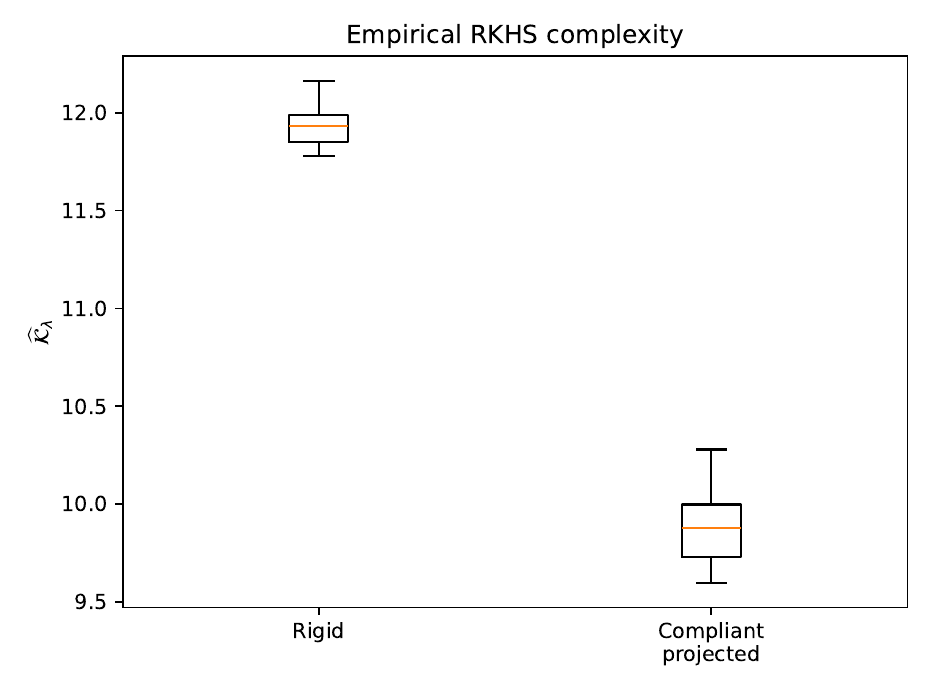}
\caption{Empirical RKHS distributional complexity over independent
realizations. The projected compliant distribution exhibits lower
feature-space spectral complexity than the rigid distribution.}
\label{fig:rkhs_complexity_comparison}
\end{figure}

Figure~\ref{fig:rkhs_complexity_comparison} reports the empirical
complexity values over the independent realizations.

\begin{table}[t]
\centering
\caption{Numerical comparison between rigid and compliant models.}
\label{tab:rkhs_numerical_results}
\begin{tabular}{lcc}
\hline
Quantity
&
Rigid
&
Compliant
\\
\hline
Projected Euclidean support volume
&
\(6.2891\)
&
\(6.2891\)
\\
Exact projected support volume
&
\(6.2832\)
&
\(6.2832\)
\\
Kernel pullback volume
&
\(5.3171\)
&
\(5.3171\)
\\
Mean task energy
&
\(0.5004\)
&
\(0.3328\)
\\
RKHS distributional complexity
&
\(11.935\)
&
\(9.865\)
\\
\hline
\end{tabular}
\end{table}

\subsubsection{Interpretation}

The numerical results separate three conceptually distinct
quantities:
\begin{itemize}
\item $ W_{\mathrm{Euc}} $ measures the coordinate volume of the viable support.
\item $ W_K $ measures the geometric volume induced by the RKHS pullback metric.
\item $ \mathcal C(P) $ measures the spectral complexity of the probability distribution
supported on that geometry.
\end{itemize}

In the present separable example, the first two quantities are the
same for the rigid and compliant projected systems because their
supports coincide. The third quantity differs because compliance
changes how the viable region is populated.

The example therefore supports the generalized interpretation
developed in the preceding subsection:

\begin{tcolorbox}[colframe=blue, colback=blue!20]
\begin{equation}
\text{geometry}
\rightarrow
W,
\ \ \ 
\text{dynamics}
\rightarrow
P,
\ \ \ 
(W,P)
\rightarrow
\mathcal C(P)
\end{equation}
\end{tcolorbox}

Accordingly, viable phase-space volume remains the zeroth-order
geometric descriptor, while RKHS distributional complexity
captures the additional organization introduced by morphology and
dynamics.

\section{Discussion}

The results provide quantitative support for a refined
interpretation of the morphological computation hypothesis.
Morphology affects locomotion not only by shaping the set of
physically viable states, but also by organizing the
probability distribution generated by the dynamics over that
set.

This distinction is particularly clear in the compliant
quadratic example. The rigid and compliant systems possess
the same projected task-space support, yet compliance
redistributes probability mass toward lower-energy task
states. The corresponding RKHS covariance spectra and
distributional complexities therefore differ despite the
equality of their projected geometric supports.

Impact mechanics provides a complementary mechanism.
Heel--strike changes both energy and coordinate
phase--space geometry, but neither the kinetic-energy
retention factor nor the Jacobian of the reset map can be
identified directly with the viable-volume ratio. Viability
depends on the subsequent hybrid dynamics.

These results suggest that the contribution of morphology
to control should be described through two complementary
objects: the geometry of the viable state space and the
probability measure induced over that geometry. Scalar
phase--space volume $W(D_v)$ remains a useful zeroth-order
descriptor of support size, whereas the RKHS formulation
captures the additional statistical organization of viable
behavior.

Within the Mind--Body Nexus framework, morphological
computation can therefore be interpreted more generally as
the capacity of physical embodiment to reshape the
metric--measure structure of viable behavior, thereby
altering the informational burden presented to the
controller.

From a more general point of view, Equation \eqref{eq:bI} relies on very general assumptions and has a wide applicability. It provides a useful method to link the system dynamics of a robot to the computational effort needed to control it. It can explain, model, and measure the benefits - and shortcomings - of `softness' in a given robot (or parts of it) for a given application, by following the approach proposed in \cite{Bonsignorio:2008:Materials}. This can be achieved by looking at how the morphology shapes the phase--space and consequently the information processing and the intelligent behaviors of a physical agent. 

Furthermore, the results presented in Section~\ref{subsec:numerical_illustration} support
a refined interpretation of the morphological computation
hypothesis.

Rather than acting solely by reducing the size of the viable
phase-space region, morphology modifies both the geometry of the
viable state space and the probability distribution induced by the
system dynamics over that space.
These two effects represent complementary mechanisms through which
the physical embodiment contributes to the overall computation
required for locomotion.
The geometric aspect is captured by the viable manifold
\(D_v\) and by its associated phase-space volume \(W\), which
quantifies the set of dynamically admissible states compatible
with successful walking.
The statistical aspect is captured by the dynamically induced
probability measure \(P\), describing how trajectories actually
populate the viable manifold.
The present work shows that these two quantities should not be confused.
Indeed, different morphologies may generate identical viable
supports while inducing substantially different probability
measures over the same state space.
Consequently, the computational burden experienced by the
controller cannot, in general, be inferred from geometric volume alone.
The RKHS formulation developed in Section~\ref{sec:rkhs} provides a natural framework for integrating these two complementary aspects.
The viable phase-space volume introduced in the first part of the paper represents a purely geometric description of locomotion complexity.
Kernel mean embeddings and covariance operators extend this description by incorporating the statistical organization of the dynamics.
The resulting RKHS distributional complexity therefore constitutes a genuine generalization of the original phase-space volume measure.
In the special case in which trajectories populate the viable
manifold uniformly, the RKHS complexity becomes a
functional of the geometry of the viable domain and of the
chosen kernel. Scalar phase--space volume remains an
important zeroth-order descriptor, but domains of equal
volume need not possess identical RKHS covariance spectra.
More generally, distributional complexity depends jointly on
the geometry of the support and on the probability measure
induced by the dynamics.
This interpretation is consistent with the Mind--Body Nexus framework.
Within this perspective, the body does not merely constrain the space of possible behaviors.
Instead, morphology actively shapes the probability structure of the behaviors generated by the interaction between body, environment, and controller.
The informational burden assigned to the controller is therefore determined not only by the size of the viable state space but also by the statistical organization of the trajectories evolving within it.
Morphological computation can thus be interpreted as an
information-geometric phenomenon emerging from the coupled effects of viable-state geometry and induced dynamics.
This viewpoint considerably broadens the applicability of the proposed framework.
The theory is not restricted to passive dynamic walking or to compliant mechanisms.
Whenever the dynamics of an embodied agent induces a probability distribution over a viable manifold, the same geometric and RKHS construction can be employed to quantify computational complexity.
Potential application domains include compliant robotic
manipulation, legged locomotion, soft robotics, adaptive
morphologies, embodied reinforcement learning, and biological sensorimotor systems.
In all these cases, the proposed framework provides a principled connection between morphology, dynamics, information processing, and controller complexity.
Several research directions naturally follow from the present work.
We emphasize that $\mathcal{C}(P)$ is an information-geometric
descriptor of the organization of viable behavior, rather than
a direct measurement of computational time, controller memory,
or control energy. The hypothesis underlying the present
framework is that reductions in the effective informational
structure presented by the embodied dynamics can reduce the
complexity required of a controller. Establishing quantitative
relations between $\mathcal{C}(P)$ and specific controller
resources constitutes an important direction for future work.
First, extending the theory to stochastic and high-dimensional systems requires efficient numerical approximations of the induced probability measure and of the associated RKHS covariance operators.
Second, the pullback geometry introduced in Section~\ref{sec:rkhs} suggests a novel class of kernel-induced metrics for comparing embodied systems that may share the same Euclidean phase-space geometry while exhibiting fundamentally different dynamical organizations.
Finally, integrating these information-geometric quantities into physics-aware learning algorithms may provide a systematic route toward morphology-aware control architectures in which learning, physical embodiment, and information processing are optimized jointly.

\section*{ACKNOWLEDGMENT}

This work has been partially supported by the PRIN project BEASTIE (2022FZJ7MR) and by ``RKHS and Applications'' (CUP C39J21042630006), both funded by the Italian Ministry of University and Research and by the European Union – NextGenerationEU, and supported by the EU project Horizon 2020 AIFORS 952275. E. Barletta, S. Dragomir and F. Esposito were partially supported by G.N.S.A.G.A - INdAM, Italy. F. Esposito is the holder of a Research Contract at the Università degli Studi della Basilicata, funded by the European Union (NEXTGENERATIONEU - CUP C43C25000280007).
 
\newpage
\makeatletter
\let\@maketitle\saved@maketitle
\makeatother
\savedtitle{\LARGE \bf Softness is Something You also Need! \\ Why Compliance Helps Control: \\ Information Metrics of a Passive Walker with Compliant Ankle \\
SUPPLEMENTAL INFORMATION
}
\savedauthor{Fabio Bonsignorio$^{1}$, Elisabetta Barletta, Sorin Dragomir, Francesco Esposito$^{2}$ and Enrica Zereik$^{3}$
}
\typeout{### COMPILING SUPPLEMENT DOCUMENT ###}
\savedmaketitle
\setcounter{section}{0}
\setcounter{equation}{0}
\setcounter{table}{0}
\setcounter{figure}{0}
\setcounter{footnote}{0}

\begin{abstract}
This document provides additional detailed information,  in particular mathematical definitions and proofs, related to the paper. They are referenced in the paper text where they were considered suitable.

\end{abstract}

\section{Calculation of \texorpdfstring{$W$}{W} Quantities}
\label{sec:wcalculation}
\subsection{How to calculate the state space volume of the solutions of a Stochastic Partial Differential Equation (SPDE)}
\setlist[description]{leftmargin=1cm,labelindent=0cm}
\begin{description}

\item[Understand the SPDE:] analyze the variables, the stochastic components (like noise terms), boundary and initial conditions, and any other relevant parameters.

\item[Numerical Simulation:] analytical solutions are often impossible to obtain.  You can use numerical methods like finite difference, finite elements, or spectral methods to discretize the SPDEs and simulate them over a range of parameters. 

\item[Monte Carlo Methods:] Monte Carlo simulations can be particularly useful. By running a large number of simulations with different realizations of the stochastic components, you can obtain a distribution of the solution space.

\item[Estimate State Space Volume:]  once you have a distribution of solutions from the Monte Carlo simulations, you can start estimating the volume of the state space by statistical techniques (e.g principal component analysis). 

\item[High--Dimensional Data Analysis:]  since the state space is likely high-dimensional, techniques from high-dimensional data analysis might be necessary. Tools like manifold learning can help in understanding the structure of the solution space.

\end{description}

The interested reader can refer to~\cite{Krugilov:2006:dimsspde} for more details.

\subsection{Usage of Reproducing Kernel Hilbert Spaces}

Reproducing Kernel Hilbert Spaces (RKHS) are an important mathematical concept and are widely used in machine learning, functional analysis, and quantum mechanics. They play a critical role in kernel methods, where they provide a way to map data into a higher-dimensional space where the data is more easily separated or manipulated.
Differential equations typically define a flow in phase--space. Evaluating the volume of a phase--space usually involves integrating over this space.

The RKHS approach provides a weighted Bergman kernel function $K_\gamma$ that maps from 
$\Omega \times \Omega$ (where $\Omega$
is the phase--space) to the set of complex numbers $\mathbb{C}$, in correspondence to an admissible weight $\gamma \in AW(\Omega )$. The 
association of $K_\gamma$ is an alternative measure to the Liouville measure, which allows the calculations of volumes based on 
complex analysis and complex geometry techniques. Renouncing real valued functions in favor of complex valued
functions (of one or several complex variables) is coherent with the philosophy in \cite{Vil:1958:complexsig} and \cite{Bouboulis:2011:wirtinger}. 
For more details see~\cite{Yildirim:2021:rkhspde} and \cite{BaDrEs:2020:axioms}.\\
\subsection{Practical Example}
\begin{wrapfigure}[10]{r}{3cm}
\includegraphics[width=2.5cm,height=6\baselineskip]{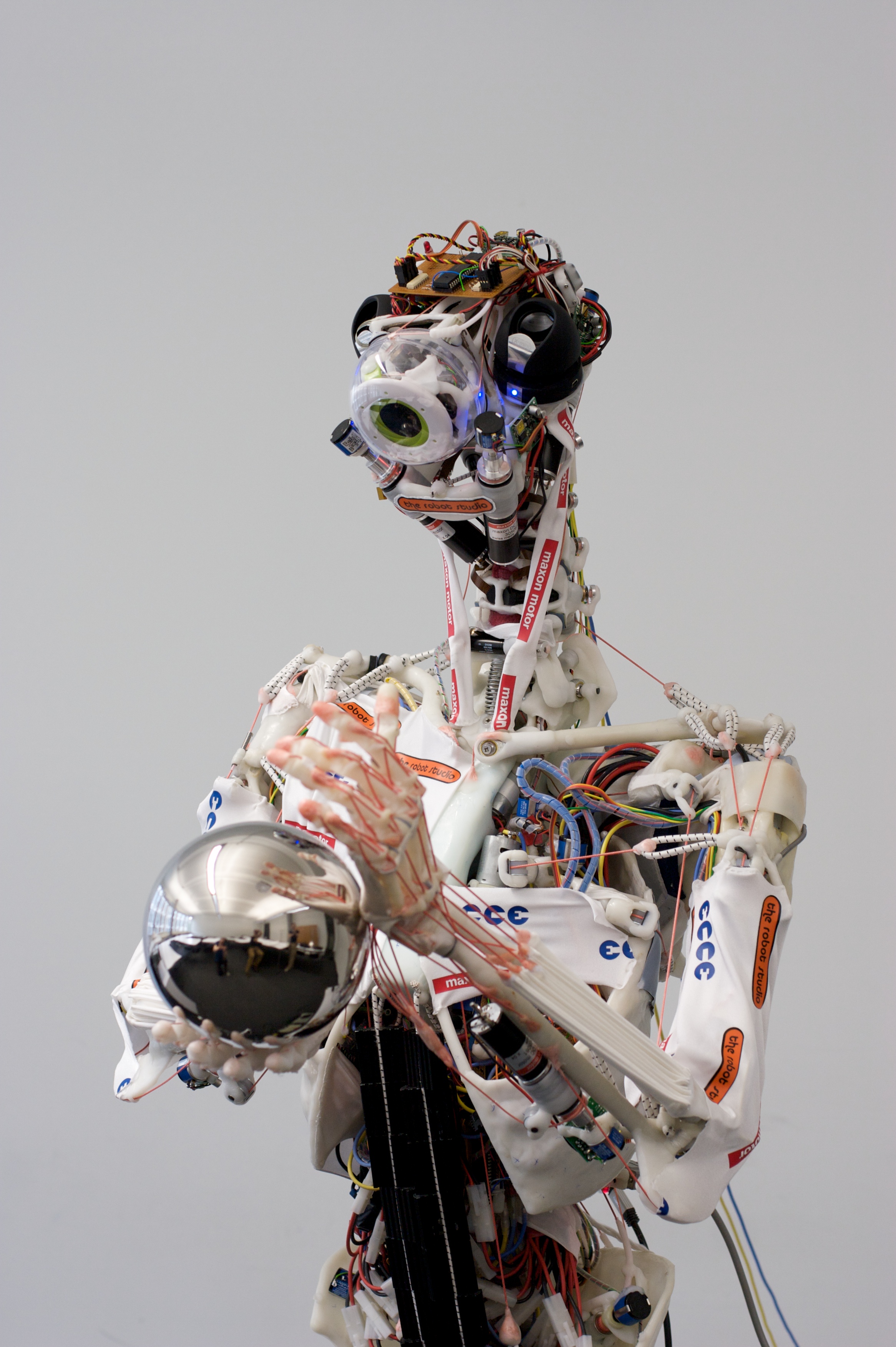}
\caption{The `Ecce' robot.}
\label{fig:Ecce}
\end{wrapfigure}
The passive walkers are not soft; however they exploit morphological computation, and they have power needs comparable to humans, as underlined in \cite{Ruina2011phisrev} and
\cite{Ruina2014IJRR}.

Why are their control programs so simple? Equation \eqref{eq:bI} in the main document allows to answer and to develop a working modeling framework  for a new breed of humanoids, like the venerable Ecce in Figure \ref{fig:Ecce}, developed in the EU project ECCEROBOT, \cite{ecceproject}. 

Calculating the phase--space volume of an inverted pendulum involves considering the variables that describe its state and the constraints they are subject to.
\begin{description}
    \item[State Variables.]  For an inverted pendulum, typical state variables include:
    \begin{itemize}
        \item \( \theta \): angle of the pendulum from the vertical (in radians);
        \item \( \dot{\theta} \): pendulum angular velocity (in radians/s).
    \end{itemize}
   \item[Phase--Space.] The phase space is spanned by the above listed variables. So, the phase--space point \( (\theta, \dot{\theta}) \) describes the pendulum state.
    \item[Constraints.] The inverted pendulum is subject to physical constraints such as conservation of energy (which determines the total energy of the pendulum) and the dynamics described by its motion equations (which relate \( \theta \) and \( \dot{\theta} \)).
 \item[Phase--Space Volume.] To calculate the phase--space volume, one needs to integrate over the accessible region in the \( (\theta, \dot{\theta}) \) space that satisfies all the relevant constraints. The volume element in phase--space is  $d \, {\rm vol} (g_B )$ where $g_B = {\mathcal K}^\ast g_{\rm Fubini-Study}$ and ${\mathcal K} : \Omega
\to \mathbb{CP} \big[ L^2 {\mathcal O} (\Omega \, , \, \gamma ) \big]$ is a coherent state map (an anti-holomorphic 
embedding). 

 \item[Energy Constraint.] The total energy \( E \) of the pendulum is typically given by \(E = \frac{1}{2} I \dot{\theta}^2 + mgh \cos(\theta)\), where \( I \) is the moment of inertia, \( m \) is the mass, \( g \) is the acceleration due to gravity, and \( h \) is the height of the pendulum's center of mass.\\
The accessible region   
$D = \big\{ (\theta \, , \, \dot{\theta} ) \in \big[ 0, \, \frac{\pi}{2} \big] \times {\mathbb R} \; : \; 
0 \leq \dot{\theta} \leq  \sqrt{\frac{2 g}{h}} \, \big( 1 - \cos \theta \big)^{1/2} \big\}$
in phase space is determined by the energy constraint \( E = \text{constant} \).
 \item[Calculating the Volume.] The phase--space volume \( V \) for a given energy \( E \) is then obtained by integrating over the allowed region in \( (\theta, \dot{\theta}) \) space:
   \[
   V(E) = \int_D \; d \; {\rm vol} (g_B ) . 
   \]

\end{description}
In summary, calculating the phase--space volume of an inverted pendulum involves setting up the energy constraint, determining the allowed region in \( (\theta, \dot{\theta}) \) space, and then integrating over that region to find the volume.

\subsection{Classical States versus Coherent States for the
Inverted Pendulum}

Equation~(1) in the main document introduces $W$ as a
geometric descriptor of the admissible phase--space region.
For simple systems such as the harmonic oscillator, this
quantity can be computed directly from the geometry of the
corresponding energy ellipse. For realistic systems, however,
the viable phase space may be curved, bounded by nonlinear
constraints, and affected by stochastic or noisy dynamics.

A more general mathematical framework is therefore useful
for defining and computing phase--space geometry. In this
section we introduce such a framework based on complex
geometry, reproducing kernel Hilbert spaces (RKHS), and
coherent states.

The purpose of this construction is to replace the intuitive
notion of ``counting classical phase--space states'' by a
continuous, geometry-aware notion of volume. Each classical
state is mapped to a coherent state in a Hilbert space using
weighted Bergman kernels. This provides a rigorous geometric
representation of the admissible phase--space support.

Importantly, the geometric volume of this support should not
be identified with the full informational complexity of the
dynamics. As developed in Section~II, morphology may modify
the geometry of the viable set, the probability measure
induced over that set, or both. The RKHS framework allows
these effects to be represented within a common mathematical
setting.

\subsubsection{Classical State}

A classical state is the usual phase--space point
\begin{equation}
z=(\theta,\dot{\theta}),
\end{equation}
which, for the inverted pendulum, corresponds physically to
the pendulum angle and angular velocity. The admissible
states form a domain
\begin{equation}
D\subset\mathbb{R}^{2}.
\end{equation}

\subsubsection{Coherent State}

Instead of treating a state only as a point in the classical
phase space, we associate each $z\in D$ with a coherent
state in a Hilbert space. Let $K_{\gamma}$ denote the
weighted Bergman kernel associated with an admissible weight
$\gamma$. The corresponding coherent-state map is
\begin{equation}
\mathcal{K}:D\rightarrow
\mathbb{CP}\!\left(L^{2}_{H}(D,\gamma)\right),
\end{equation}
with
\begin{equation}
\mathcal{K}(z)=[K_{\gamma}(\cdot,z)].
\end{equation}
Thus, each classical state is represented by a ray in a
projective Hilbert space.

\subsubsection{Why Use Coherent States?}

Direct computation of volumes in nonlinear or curved
phase--space domains can be difficult. The coherent-state
construction equips the classical domain with a geometry
induced by the RKHS representation.

In this representation, geometric quantities can be obtained
from kernel functions and their associated metric and
curvature structures. The basic strategy is therefore
\begin{equation}
\text{classical state}
\longrightarrow
\text{coherent state}
\longrightarrow
\text{kernel-induced geometry}.
\end{equation}

This construction provides a geometry-aware representation
of the viable state space and complements direct
phase--space integration.

\subsubsection{Overview of the Construction}

The construction proceeds through the following steps.

First, define a weighted Hilbert space on the domain $D$.
Let
\begin{equation}
\gamma:D\rightarrow[0,+\infty)
\end{equation}
be an admissible weight and let $L^{2}_{H}(D,\gamma)$
denote the corresponding Hilbert space of square-integrable
holomorphic functions. Its reproducing kernel is denoted by
$K_{\gamma}(\zeta,z)$.

Second, construct the coherent-state map
\begin{equation}
\mathcal{K}:D\rightarrow
\mathbb{CP}\!\left(L^{2}_{H}(D,\gamma)\right).
\end{equation}

Third, the overlap between two coherent states provides the
transition amplitude
\begin{equation}
a^{00}(\zeta,z)
=
\left\langle
\mathcal{K}(z),\mathcal{K}(\zeta)
\right\rangle .
\end{equation}

Finally, the Liouville volume of the admissible domain can
be related to the diagonal of an appropriately weighted
Bergman kernel. For the family of weights introduced below,
\begin{equation}
\operatorname{vol}_{L}(D)
=
\int_D d\mu_L
=
\frac{C(s)}{2}
\int_D
K_{\gamma_s}(\zeta,\zeta)
\gamma_s(\zeta)\,
d\zeta\wedge d\bar{\zeta}.
\label{eq:bergman-volume}
\end{equation}
Thus, the geometrical quantity $W(D)$ can be evaluated
through the kernel representation of the domain rather than
only through direct coordinate-space integration.

\subsubsection{Relation to the Mind--Body Nexus}

The geometric construction above provides one component of
the Mind--Body Nexus framework. Morphology determines, or
contributes to determining, the viable geometry accessible to
the physical system. The hybrid dynamics additionally induce
a probability measure over that geometry.

The resulting hierarchy is therefore
\begin{equation}
\resizebox{0.95\columnwidth}{!}{$%
\boxed{
\mathrm{Morphology}
\longrightarrow
\bigl(D_v,W(D_v)\bigr)
\longrightarrow
P
\longrightarrow
(\mu_P,C_P)
\longrightarrow
\mathcal{C}(P)
}
$}%
\label{eq:mind-body-hierarchy}
\end{equation}
where $D_v$ denotes the viable support and $W(D_v)$ is a
zeroth-order descriptor of its geometric size. The
probability measure $P$ describes how the dynamics populate
that support, while $\mu_P$ and $C_P$ denote, respectively,
its RKHS mean embedding and centered covariance operator.

The corresponding distributional complexity is
\begin{equation}
\mathcal{C}(P)
=
\log\det\nolimits_F
\left(I+\gamma C_P\right).
\end{equation}

Consequently, morphological computation should not be
identified exclusively with phase--space contraction.
Morphology may alter the geometry of the viable set, modify
the probability distribution induced over an unchanged
support, or produce both effects. The relevant object is
therefore more generally the metric--measure structure
$(D_v,P)$ rather than the scalar volume $W(D_v)$ alone.

This distinction provides the formal bridge between body
mechanics and information-processing complexity used in the
main paper and developed further in Section~II of this
Supplemental Information.

\subsubsection{Formal Development}

We now provide the mathematical details of the construction.
A weight function is a measurable map
\begin{equation}
\gamma:D\rightarrow[0,+\infty).
\end{equation}
Let $L^{2}(D,\gamma)$ be the Hilbert space of square
integrable functions $f:D\rightarrow\mathbb{C}$ satisfying
\begin{equation}
\int_D |f(z)|^{2}\gamma(z)\,d\mu(z)<\infty,
\end{equation}
with inner product
\begin{equation}
(f,g)_{\gamma}
=
\int_D
f(z)\overline{g(z)}
\gamma(z)\,d\mu(z),
\end{equation}
where $\mu$ denotes Lebesgue measure on $\mathbb{R}^{2}$.

Let $L^{2}_{H}(D,\gamma)$ denote the subspace of
holomorphic functions in $L^{2}(D,\gamma)$. A weight
$\gamma$ is admissible if the evaluation functional
\begin{equation}
\delta_z:L^{2}_{H}(D,\gamma)\rightarrow\mathbb{C},
\qquad
\delta_z(f)=f(z),
\end{equation}
is continuous and $L^{2}_{H}(D,\gamma)$ is a closed
subspace of $L^{2}(D,\gamma)$.

Under these assumptions the evaluation functional is
represented by the weighted Bergman kernel
$K_{\gamma}(\zeta,z)$, and $L^{2}_{H}(D,\gamma)$ is an
RKHS.

The coherent-state map is then
\begin{equation}
\mathcal{K}:D\rightarrow
\mathbb{CP}\!\left(L^{2}_{H}(D,\gamma)\right),
\qquad
z\mapsto[K_{\gamma}(\cdot,z)].
\end{equation}

For classical states $z,\zeta\in D$, the associated
transition probability amplitude is
\begin{equation}
a^{00}(\zeta,z)
=
\left\langle
\mathcal{K}(z),\mathcal{K}(\zeta)
\right\rangle .
\end{equation}

Let
\begin{equation}
\mathbb{B}=\{z\in\mathbb{C}:|z|<1\}
\end{equation}
be the unit disk and let
\begin{equation}
R:\mathbb{B}\rightarrow D
\end{equation}
be a Riemann map. For $s>-1$, define
\begin{equation}
\phi_s(\zeta)
=
\left(1-|\zeta|^{2}\right)^s,
\end{equation}
and obtain an admissible weight on $D$ by pullback,
\begin{equation}
\gamma_s
=
\phi_s\circ R^{-1}.
\end{equation}

The corresponding weighted Bergman kernel transforms under
the conformal map according to the appropriate kernel
transformation law. With the normalization adopted here, the
Liouville volume can then be written in the form of
Eq.~\eqref{eq:bergman-volume}.

The central point for the present paper is that this
construction provides a rigorous geometric representation of
the viable domain $D_v$ and its associated volume
$W(D_v)$. The probability measure $P$ induced by the
physical dynamics is a distinct object. Section~II makes this
distinction explicit and introduces the corresponding
distributional RKHS description.

\subsection{Application Example}
The approach described in the preceding subsection is quite
general. We illustrate it here using the inverted pendulum,
for which less abstract approaches are also available.
We detail the computational procedure step by step:

\subsubsection{Map the domain \texorpdfstring{$D$}{D} to the unit disk}
since $D$ is simply connected (no holes), there exists a conformal (Riemann) map
$R: \mathbb{B}\to D$
from the unit disk $\mathbb{B}=\{z:|z|<1\}$ to the domain $D$.  

\subsubsection{ Choose a weight on the disk and pull it back to \texorpdfstring{$D$}{D}}
we set a family of weights on the disk,
$\phi_s(\zeta) = (1-|\zeta|^2)^s$, 
then we define a weight on $D$ by pulling back through $R$:
$\gamma_s = \phi_s\circ R^{-1}$.

\subsubsection{ Compute the weighted Bergman kernel on  \texorpdfstring{$D$}{D}}
we start from the known kernel on the disk (closed form), then transform it via the mapping $R$. The transformation rule is
$K_{\gamma_s}(\zeta,z)
= R'(\zeta)\,\overline{R'(z)}\; K_{\phi_s}(R(\zeta),R(z))$.
Conceptually, the derivative terms correct for geometric distortion under the map. 

\subsubsection{Get volume from the kernel diagonal}
the key computational shortcut is
based on the expression $\mathrm{vol}(D)\ \propto\ \int_D K_{\gamma_s}(z,z)\,\gamma_s(z)\,dz$, which holds up to constants like $C(s)$ above and the chosen measure conventions.  

\subsubsection{Practical approximation (what you actually implement)} 
since $R$ is rarely analytically known, you will need to:
\begin{itemize}
    \item approximate $D$ by polygons $P_N$
    \item compute polygon-to-disk maps via Schwarz–Christoffel methods (through one of the mentioned existing fast algorithms)
    \item get approximations $R_N \to R$
    \item build $\gamma_{s,N}$ and $K_{\gamma_{s,N}}$
    \item integrate numerically to retrieve the volume.
\end{itemize}

{\small {\bf Historical Note}. Bergman kernels (unweighted, so that $\gamma \equiv 1$) are due to Steven Bergman \cite{Ber}, and an early account of their
wide applicability is furnished in the monograph \cite{BeSc}. In 1952
Gaetano Fichera started a personal crusade (refer to \cite{Fic}) against Bergman kernels, on the ground of their being ``inessential'' {\em vis-a-vis} to other methods in Mathematical Analysis. Bergman kernels fit into the general theory of RKHS \cite{Aro}, which was already available at the time Fichera's paper \cite{Fic} was written. 
The further growth of the use of kernel methods in the theory of 
PDEs (discussed e.g. in \cite{Lo} and \cite{Wei}), complex geometry (relative to Bergman metrics, see pages 367-371 of \cite{Hel}) and approximation theory (in, e.g., \cite{BaDr} and \cite{BBDEZ}) suggest that the great Italian scientist was, at least once, abandoned by his unquestionable  mathematical intuition.}

\section{Physical Interpretation: Support Geometry and
Distributional Complexity}
\label{PhysicalInterpr}
Compliance introduces internal energy-storage degrees of freedom
that modify how the available mechanical energy is distributed
between task variables and internal morphological variables.
It is important, however, to distinguish the geometric support
of the admissible task states from the probability measure
induced on that support.

Let
\begin{equation}
z=(x,\eta),
\end{equation}
where $x$ denotes the task variables and $\eta$ the compliant
internal degrees of freedom. For analytical tractability, assume
the separable Hamiltonian
\begin{equation}
H(z)=H_x(x)+H_\eta(\eta),
\qquad H_\eta(\eta)\geq 0,
\end{equation}
and define
\begin{align}
D_r(E)
&=
\{x:H_x(x)\leq E\},\\
D_c(E)
&=
\{(x,\eta):H_x(x)+H_\eta(\eta)\leq E\}.
\end{align}

\subsection{Projection of the compliant energy domain}

\textbf{Proposition II.1 (Projected-support equivalence).}
Assume that the compliant subsystem possesses an admissible
zero-energy state $\eta_0$ such that
\begin{equation}
H_\eta(\eta_0)=0.
\end{equation}
Then
\begin{equation}
\Pi_x D_c(E)=D_r(E).
\label{eq:projected_support_equality}
\end{equation}

\emph{Proof.}
If $x\in\Pi_xD_c(E)$, there exists an $\eta$ such that
\[
H_x(x)+H_\eta(\eta)\leq E.
\]
Since $H_\eta(\eta)\geq0$, it follows that $H_x(x)\leq E$,
and therefore
\[
\Pi_xD_c(E)\subseteq D_r(E).
\]
Conversely, let $x\in D_r(E)$. Choosing $\eta=\eta_0$ gives
\[
H_x(x)+H_\eta(\eta_0)=H_x(x)\leq E,
\]
and hence $(x,\eta_0)\in D_c(E)$. Therefore
\[
D_r(E)\subseteq\Pi_xD_c(E).
\]
The two inclusions prove Eq.~\eqref{eq:projected_support_equality}.
\hfill$\square$

Thus, for the separable model considered here, compliance does
not by itself contract the projected geometric support of the
task variables. Its effect appears instead in the measure induced
on that support, because the additional compliant coordinates
redistribute the available energy.

\subsection{Energy redistribution under uniform sampling}

Consider the local quadratic model
\begin{equation}
H_x=
\frac{1}{2}a\theta^2+
\frac{1}{2}b\omega^2,
\qquad
H_\eta=
\frac{1}{2}k\phi^2+
\frac{1}{2}J\dot{\phi}^{\,2}.
\label{eq:quadratic_compliant_model}
\end{equation}

After rescaling the four coordinates, the compliant energy
domain becomes a four-dimensional Euclidean ball,
\begin{equation}
u_1^2+u_2^2+u_3^2+u_4^2\leq 2E.
\end{equation}
For a point sampled uniformly in a $d$-dimensional ball of
radius $R$,
\begin{equation}
\mathbb{E}[r^2]=\frac{d}{d+2}R^2.
\end{equation}
For $d=4$ and $R^2=2E$, this gives
\begin{equation}
\mathbb{E}[r^2]=\frac{4}{6}(2E)=\frac{4E}{3}.
\end{equation}

By rotational symmetry, each quadratic coordinate contributes
one fourth of this expectation. Since $H_x$ contains two of
the four quadratic coordinates,
\begin{equation}
\boxed{
\mathbb{E}[H_x]_{\rm compliant}=\frac{E}{3}.
}
\label{eq:mean_task_energy_compliant}
\end{equation}

For comparison, uniform sampling inside the two-dimensional
rigid energy ellipse gives
\begin{equation}
\boxed{
\mathbb{E}[H_x]_{\rm rigid}=\frac{E}{2}.
}
\label{eq:mean_task_energy_rigid}
\end{equation}

Therefore,
\begin{equation}
\mathbb{E}[H_x]_{\rm compliant}
<
\mathbb{E}[H_x]_{\rm rigid},
\end{equation}
even though
\begin{equation}
\Pi_xD_c(E)=D_r(E).
\end{equation}

The effect of compliance in this separable example is therefore
not a contraction of the projected support, but a redistribution
of probability mass toward lower-energy task states. This
distinction motivates the distributional RKHS complexity
measure introduced in the main paper.

\subsection{Distributional complexity}

Let $P_r$ and $P_c$ denote the probability measures induced
on the task space by the rigid and compliant systems,
respectively. Although their supports may coincide,
\begin{equation}
\operatorname{supp}(P_r)
=
\operatorname{supp}(P_c)
=
D_r(E),
\end{equation}
the measures need not coincide:
\begin{equation}
P_r\neq P_c.
\end{equation}

Let $C_P$ denote the centered covariance operator associated
with the RKHS embedding of $P$, and define
\begin{equation}
\mathcal{C}(P)
=
\log\det_F\!\left(I+\gamma C_P\right).
\end{equation}

If
\begin{equation}
0\preceq C_{P_c}\preceq C_{P_r},
\end{equation}
then, by monotonicity of the eigenvalues of positive
trace-class operators,
\begin{equation}
\mathcal{C}(P_c)
\leq
\mathcal{C}(P_r).
\label{eq:rkhs_complexity_ordering}
\end{equation}

This result makes explicit that the reduction of informational
complexity is a property of the induced probability
distribution, rather than a necessary consequence of a
reduction in scalar phase--space volume.

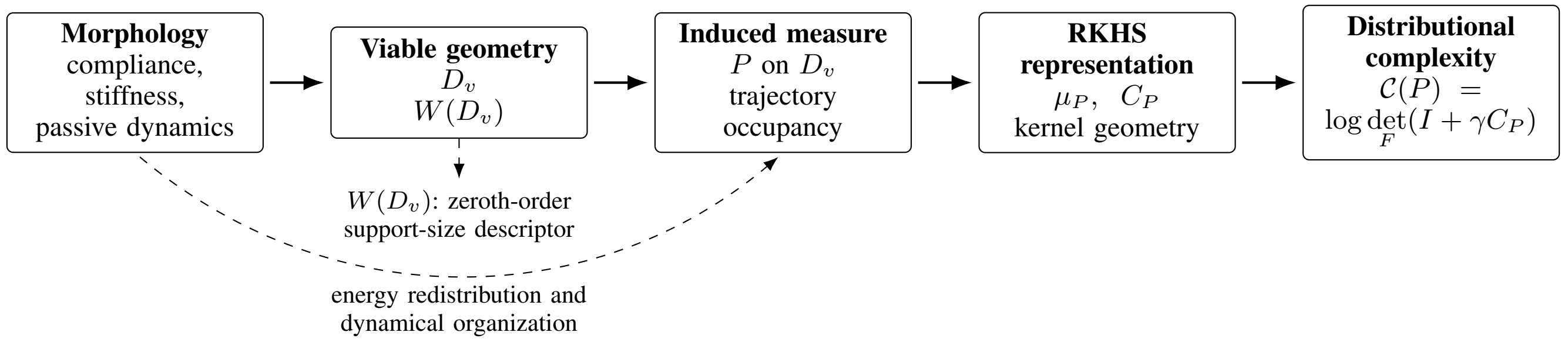
\begin{figure*}[t]
\centering

\begin{tikzpicture}[
    node distance=0.75cm,
    >=Latex,
    box/.style={
        draw,
        rounded corners=2pt,
        align=center,
        minimum height=1.25cm,
        text width=2.65cm,
        font=\small
    },
    arrow/.style={
        ->,
        thick,
        shorten >=2pt,
        shorten <=2pt
    }
]

\node[box] (morph)
{\textbf{Morphology}\\
compliance, stiffness,\\
passive dynamics};

\node[box, right=of morph] (geometry)
{\textbf{Viable geometry}\\
$D_v$\\
$W(D_v)$};

\node[box, right=of geometry] (measure)
{\textbf{Induced measure}\\
$P$ on $D_v$\\
trajectory occupancy};

\node[box, right=of measure] (rkhs)
{\textbf{RKHS representation}\\
$\mu_P,\;C_P$\\
kernel geometry};

\node[box, right=of rkhs] (complexity)
{\textbf{Distributional complexity}\\
$\displaystyle
\mathcal{C}(P)
=
\log\det_F(I+\gamma C_P)$};

\draw[arrow] (morph) -- (geometry);
\draw[arrow] (geometry) -- (measure);
\draw[arrow] (measure) -- (rkhs);
\draw[arrow] (rkhs) -- (complexity);

\node[
    align=center,
    font=\footnotesize,
    below=.45cm of geometry
] (volume)
{$W(D_v)$: zeroth-order\\support-size descriptor};

\draw[->, dashed] (geometry) -- (volume);

\draw[
    ->,
    dashed,
    bend right=40,
    shorten >=2pt,
    shorten <=2pt
]
(morph.south) to
node[below, align=center, font=\footnotesize]
{energy redistribution and\\dynamical organization}
(measure.south);

\end{tikzpicture}

\caption{
Hierarchy connecting morphology to information-processing
complexity. Morphology determines the geometry of the viable
state space $D_v$ and can simultaneously modify the probability
measure $P$ induced by the dynamics over that space. The scalar
volume $W(D_v)$ provides a zeroth-order descriptor of support
size, whereas the RKHS mean embedding $\mu_P$ and covariance
operator $C_P$ retain information about the statistical
organization of the viable dynamics. The resulting
distributional complexity
$\mathcal{C}(P)=\log\det_F(I+\gamma C_P)$ therefore depends on
the metric--measure structure of the viable dynamics rather
than on scalar phase--space volume alone.
}
\label{fig:morphology-rkhs-complexity}

\end{figure*}

Figure~\ref{fig:morphology-rkhs-complexity} summarizes the
resulting hierarchy. In particular, morphological computation
should not be identified exclusively with contraction of the
viable support. Morphology may modify either the geometry of
the viable set, the probability measure induced over that set,
or both. The RKHS formulation provides a common representation
of these effects.

\section{A Simple Numerical Estimation of Phase--Space Volume and Complexity}
\label{appendix:simplnumest}

The viable phase--space volume $W$ is estimated using a Monte--Carlo sampling procedure.

\subsection{Sampling Procedure}

States are sampled uniformly inside a bounding box $B$ in the full state space:

\begin{equation}
z_i \sim \mathcal{U}(B).
\end{equation}

Each sample is accepted if it satisfies

\begin{equation}
H(z_i) \le E,
\end{equation}

and, for walking models, if the heel--strike reset produces a valid post--impact state.

The volume is estimated as:

\begin{equation}
W \approx \mathrm{Vol}(B) \frac{N_{\mathrm{acc}}}{N}.
\end{equation}

\subsection{Kernel Complexity Estimation}

Let $k(\cdot,\cdot)$ denote a positive-definite kernel.
In the numerical examples we employ the Gaussian kernel

\[
k(x_i,x_j)
=
\exp\!\left(
-\frac{\|x_i-x_j\|^2}{2\sigma^2}
\right).
\]

The corresponding Gram matrix is

\[
G_{ij}=k(x_i,x_j).
\]

To avoid numerical singularities, a small regularization
parameter $\lambda>0$ is introduced and the complexity
proxy is defined as: 

\begin{equation}
K(X) = \log \det(G + \lambda I),
\end{equation}

which we use here as a finite-sample kernel complexity
proxy. It should be distinguished from the operator-level
distributional complexity
$\mathcal{C}(P)=\log\det_F(I+\gamma C_P)$ introduced in
Section~II and in the main paper. The latter is defined from
the centered RKHS covariance operator of the induced
probability measure, whereas $K(X)$ is a numerical
finite-sample descriptor computed directly from the Gram
matrix.

\subsection{Effect of Compliance}

Compliance introduces additional energy-storage degrees of
freedom and can therefore modify the probability distribution
induced over the task variables. Importantly, this need not
imply a contraction of the projected task-space support. For
the separable quadratic model of Section II, the projected
rigid and compliant supports coincide,
\begin{equation}
\Pi_x D_c(E)=D_r(E),
\end{equation}
while the corresponding probability measures differ.

Consequently, changes in RKHS distributional complexity
should be attributed to changes in the induced measure and
its covariance spectrum, rather than, in general, to a
reduction of the scalar projected phase--space volume.

\section{Full State--Dependent Monte--Carlo Estimation}
\label{sec:montecarlo}

\subsection{Goal}

Estimate the measure/volume of the post--impact viable set $D^+(\gamma)$ (a subset of state space right after heel-strike) such that the walker can successfully complete the next step on slope $\gamma$.

\subsection{Post-impact viable set \texorpdfstring{$D^+$}{D+}}

Right after heel-strike (the reset), we define the post-impact state:
\begin{equation}
x^+ \;=\; \begin{bmatrix} q^+ \\ \dot q^+ \end{bmatrix}
\quad \text{\small (or higher-dimensional if you include more joints).}
\end{equation}

We assume:
\begin{enumerate}
	\item 	Post-impact configuration uncertainty / tolerance
\begin{equation}
q^+ \in [-\alpha, +\alpha] \pm dq
\end{equation}
meaning: the post-impact configuration lies in a narrow band around the nominal reset configuration (for a rimless/compass-like step, nominal is typically $q^+\approx -\alpha$, we further assume a symmetric band).
	\item 	Post-impact energy bound
\begin{equation}
T^+ \le E^+(\gamma)
\end{equation}
where $T^+$ is post-impact kinetic energy (e.g. $T^+ = \tfrac12 I(\dot q^+)^2$ in 1-DOF), and $E^+(\gamma)$ is an available energy budget that can depend on slope (or actuation limits, etc.). In practice this gives an admissible range for velocities:
\begin{equation}
|\dot q^+| \le \dot q_{\max}(\gamma)
\quad \text{with}\quad
\dot q_{\max}(\gamma)=\sqrt{\frac{2E^+(\gamma)}{I}}.
\end{equation}
\end{enumerate}

\balance

\subsection{Monte-Carlo acceptance}

We want the fraction of post-impact states in that admissible box that are actually viable.

Algorithm (state-dependent, because acceptance depends on dynamics and reset):
\begin{enumerate}
	\item Sample $N$  post-impact states $x_k^+$ uniformly (or quasi-uniformly) from:
    \begin{itemize}
	\item $q^+ \in [q_0-dq,\, q_0+dq]$ (or your symmetric interval)
	\item $\dot q^+ \in [-\dot q_{\max}(\gamma),\, +\dot q_{\max}(\gamma)]$
or only $\dot q^+>0$ if physically required.
  \end{itemize}
	\item 	Propagate each sampled state forward through the continuous dynamics until the next heel-strike event.
	\item 	Apply acceptance test:
    \begin{itemize}
     
	 \item 	“Success” if the trajectory reaches the next heel-strike event without  failure (no reversal, no fall, no forbidden constraint violation, etc.).
 \item Optionally include additional constraints (foot clearance, impact velocity bounds, etc.).
    \end{itemize}
    \item 	Estimate viable phase-space volume.
If your sampling region has known volume $\mathrm{Vol}(\mathcal{B})$, then:
\begin{equation}
\widehat{W}(\gamma)=\mathrm{Vol}(\mathcal{B})\;\frac{N_{\text{acc}}}{N}
\end{equation}
\end{enumerate}
The variance comes from a binomial estimator; you can attach confidence intervals if you want:
\begin{equation}
\text{SE}\approx \mathrm{Vol}(\mathcal{B})\sqrt{\frac{\hat p(1-\hat p)}{N}}
\quad \text{with } \hat p=N_{\text{acc}}/N
\end{equation}

\subsection{“Exact heel-strike reset”}

This means that when the next impact event happens, you use the exact discrete reset map:
\begin{equation}
x^+_{k+1}= \Delta\!\left(x^-_{k+1}\right)
\end{equation}
rather than an approximate one. 

\subsection{Exact heel--strike contraction in the reduced model}

For the simplest passive walker considered in the main paper,
the plastic heel--strike law is
\begin{equation}
\omega^+
=
\cos(2\alpha)\,\omega^- .
\label{eq:supp_impact_reset}
\end{equation}

For $\alpha=0.3\,\mathrm{rad}$,
\begin{equation}
\rho
=
\cos(2\alpha)
=
\cos(0.6)
\simeq 0.825336 .
\end{equation}

The kinetic energy associated with the angular velocity
therefore changes according to
\begin{equation}
\frac{T^+}{T^-}
=
\frac{\frac{1}{2}I(\omega^+)^2}
     {\frac{1}{2}I(\omega^-)^2}
=
\cos^2(2\alpha)
\simeq 0.681179 .
\label{eq:supp_impact_energy}
\end{equation}

Hence approximately $31.9\%$ of the corresponding pre-impact
kinetic energy is dissipated by the plastic collision.

This energetic contraction must be distinguished from the
geometric contraction of phase--space area. In the reduced
state $x=(\theta,\omega)$, leg relabeling has unit absolute
Jacobian, while the velocity component is multiplied by
$\cos(2\alpha)$. Consequently,
\begin{equation}
\left|\det D\Delta\right|
=
|\cos(2\alpha)|
\simeq 0.825336 .
\label{eq:supp_impact_jacobian}
\end{equation}

Thus the reset map alone contracts coordinate phase--space
area by approximately $17.5\%$, whereas the kinetic-energy
retention factor is approximately $0.681$. Neither quantity
should be identified directly with the ratio of viable
phase--space volumes.

The latter additionally depends on the post-impact viability
conditions. If $D^-$ denotes a pre-impact domain, the
post-impact viable set may be written as
\begin{equation}
D_v^+
=
\left\{
\Delta(x^-):
x^-\in D^-,
\quad
\Delta(x^-)\ 
\right\}.
\end{equation}
Its volume must therefore be determined from the full hybrid
dynamics, for example through the state-dependent
Monte--Carlo procedure described above.

\bibliographystyle{IEEEtran} 
\bibliography{biblioMindBody2026}
\end{document}